\documentclass{article}
\usepackage{iclr2027_conference,times}
\usepackage{amsmath,amsfonts,bm}

\def\eqref#1{equation~\ref{#1}}
\def\1{\bm{1}}

\DeclareMathAlphabet{\mathsfit}{\encodingdefault}{\sfdefault}{m}{sl}
\SetMathAlphabet{\mathsfit}{bold}{\encodingdefault}{\sfdefault}{bx}{n}

\usepackage[utf8]{inputenc}
\usepackage[T1]{fontenc}
\usepackage{textcomp}
\usepackage{newunicodechar}
\usepackage{algorithm}
\usepackage{algpseudocode}
\newunicodechar{Δ}{\ensuremath{\Delta}}
\newunicodechar{−}{\ensuremath{-}}
\newunicodechar{→}{\ensuremath{\rightarrow}}
\newunicodechar{∩}{\ensuremath{\cap}}
\newunicodechar{≥}{\ensuremath{\geq}}
\newunicodechar{≤}{\ensuremath{\leq}}
\newunicodechar{²}{\ensuremath{^{2}}}
\usepackage{hyperref}
\usepackage{url}
\usepackage{graphicx}
\usepackage{booktabs,longtable,array,calc,multirow,tabularx,makecell,adjustbox}
\usepackage{xcolor}
\usepackage{wrapfig}
\usepackage{placeins}\usepackage{flafter}  % floats never appear before their place in the text; \FloatBarrier ends each appendix
\providecommand{\tightlist}{\setlength{\itemsep}{0pt}\setlength{\parskip}{0pt}}
\title{%
  \centering
  \normalfont\bfseries
  \fontsize{17}{20}\selectfont
  MedSkill-E$^{2}$: Evaluating and Evolving Skills\\
  for Medical Agents\par
}

\author{%
\parbox{0.95\textwidth}{%
\centering
\normalfont\small\bfseries
Jingjing Hua$^{1,2}$,
Shaocong Wei$^{1,2,*}$,
Jianghan Shen$^{3}$,
Junzhi Ning$^{1}$\\
Min Cen$^{1}$,
Cheng Tang$^{2}$,
Ziyan Huang$^{1}$,
Jiepeng Li$^{2}$,
Fengshuo Liu$^{4}$\\
Kangkun Mao$^{1}$,
Mouxiao Bian$^{1}$,
Yirong Chen$^{1}$,
Ming Hu$^{1}$,
Jiyao Liu$^{5,*}$\\[8pt]
\normalfont\footnotesize
$^{1}$ Shanghai Artificial Intelligence Laboratory\\
$^{2}$ University of Chinese Academy of Sciences\\
$^{3}$ Nanjing University\\
$^{4}$ Imperial College London\\
$^{5}$ Fudan University\\[5pt]
$^{*}$ \texttt{weishaocong@pjlab.org.cn}\quad
\texttt{liujy22@m.fudan.edu.cn}
}%
}

\iclrfinalcopy

\iclrfinalcopy

\begin{document}
\raggedbottom  
\pagestyle{plain}
\maketitle
\thispagestyle{plain}

\begin{abstract}
% generated by build_en_fable.py from ../en_fable_md/0_abstract.md -- do not edit by hand
Large language model (LLM) agents increasingly use skills to access domain knowledge and complete real-world tasks. We observe that even when a skill specifies the rules and procedures needed to complete a task, agents may still fail to apply them effectively during autonomous execution, a phenomenon we call skill Enable But Underuse (EBU). This issue is particularly consequential in medical tasks, where overlooking even a single critical constraint can lead to a plausible but incorrect result. To systematically evaluate and address this phenomenon, we introduce MedSkill-E², a framework for evaluating and evolving skills for medical agents. We construct EBU-Bench, a challenging benchmark covering 12 medically relevant task groups and their associated skills, to evaluate agents' ability to use available skills autonomously. We quantify the EBU gap as the difference in skill-use accuracy between autonomous use and explicitly guided reading and application of the same skill. On EBU-Bench, all evaluated model--harness combinations exhibit EBU, with a mean gap of 43.94 percentage points (pp). For the same model, skill-use accuracy under autonomous use differs by up to 38.96 pp across harnesses. Motivated by these observations, we propose EBU-Evolve, which uses execution feedback from autonomous and explicitly guided skill use to distill effective guidance into reusable skill instructions. Relative to their pre-evolution versions, the revised skills improve skill-use accuracy during autonomous execution by 33.33--55.17 pp and continue to yield improvements when transferred across multiple models and harnesses. Our results show that skill-use accuracy under autonomous execution varies substantially across harnesses even when the model and skill are held fixed, and model rankings also change; EBU can persist even after a skill is opened. EBU-Evolve distills effective guidance into reusable skill instructions, improving skill-use accuracy under autonomous execution on unseen tasks within the same skill, and still yields improvements when the evolved skills are transferred to other models and harnesses.

\end{abstract}

% generated by build_en_fable.py from ../en_fable_md/1_introduction.md -- do not edit by hand
\section{Introduction}

Large language model (LLM) agents use skills to access domain knowledge and apply it to specific tasks. A skill organizes the methods, tool-use instructions, and operating procedures needed for a class of tasks into a reusable resource, so that agents need not work out a solution procedure anew for each problem. This support is particularly important in medicine and related research: clinical calculations require accurate application of formulas and their constraints, while evidence review requires adherence to explicit assessment criteria. Skills offer a direct way to standardize these operations. Skills have also begun to emerge for healthcare and the life sciences \citep{xu2026empiricalstudyagentskills}.

\begin{figure}[!t]
  \centering
  \includegraphics[width=\textwidth]{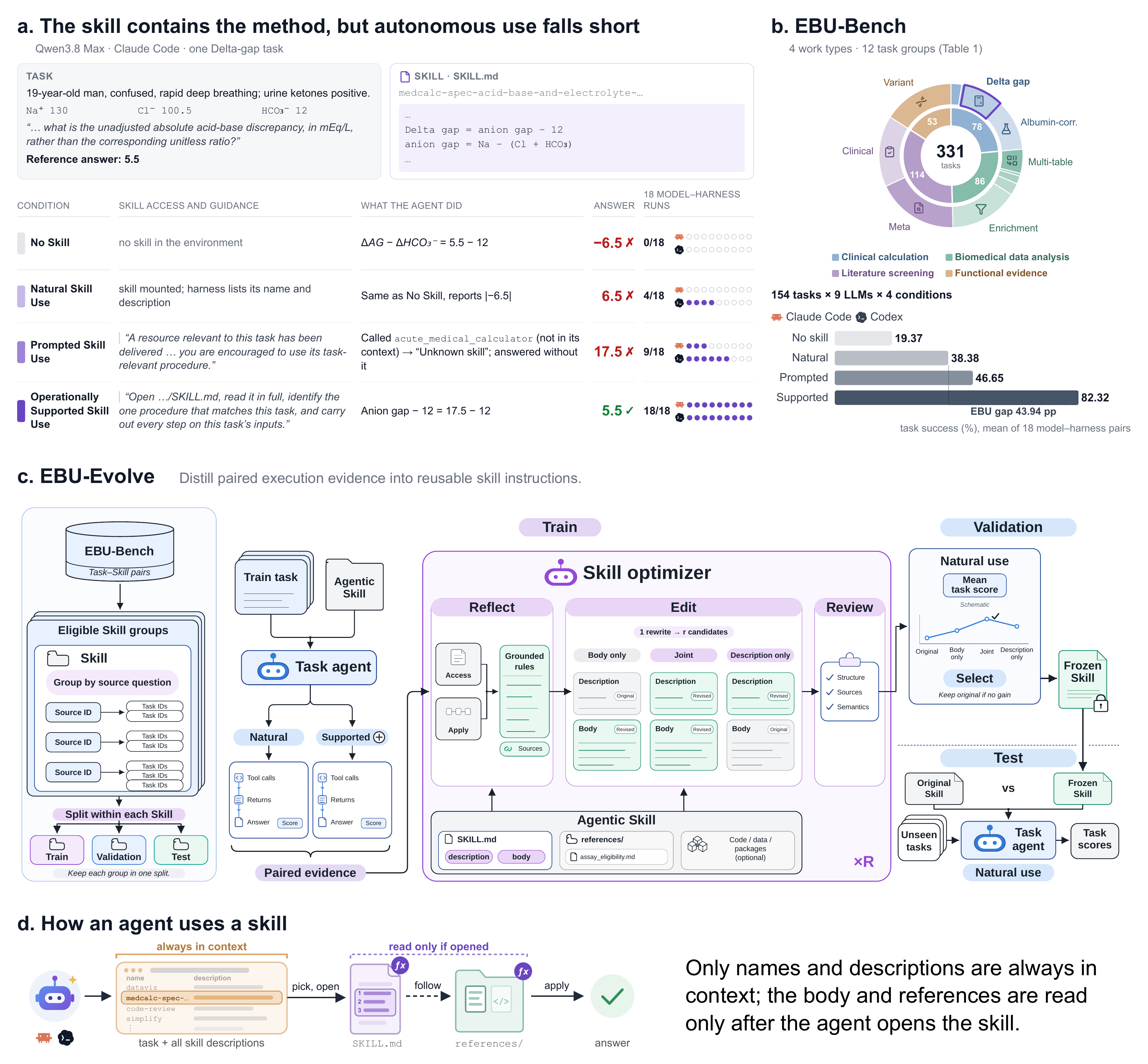}
  \caption{An example of skill underuse, the benchmark, and the evolution method. (a) Answers and behavior on the same task under four conditions; the dots on the right summarize the number of correct answers by harness, and their positions do not correspond to model order. (b) Task composition and mean skill-use accuracy under each condition on EBU-Bench Lite. (c) EBU-Evolve uses intervention-to-instruction distillation to convert same-task execution differences into reusable instructions grounded in the original materials. One edit yields body-only, joint, and description-only candidates. Reviewed candidates undergo Natural Skill Use validation; absent improvement, the original is retained. The selected skill is frozen for held-out testing. $R$ denotes independent optimization runs. (d) How an agent uses a skill.}
  \label{fig:overview}
\end{figure}

However, on medical benchmarks and practical medical research tasks, we observe that agents sometimes fail to apply the methods in a skill effectively, even when an applicable skill is provided and its purpose and conditions of use are specified in its description. In other cases, changing the execution framework (harness) while holding the model and skill fixed reduces or even eliminates the performance gains previously obtained from the skill. These observations raise a question: in real medical applications, how much assistance can an available skill provide, and how much of that benefit can an agent realize when solving tasks autonomously? We further find that some previously unsuccessful tasks can be completed when the agent is directed to use a particular skill and given corresponding execution requirements. We call this additional guidance ``operational support'' and use task performance under this condition as an empirical reference for the benefits achievable with the skill. We term the phenomenon in which the same skill supports task completion under operational support but its corresponding benefits remain unrealized during autonomous execution skill Enable But Underuse (EBU). EBU may not be widespread across medical tasks, but when it occurs, agents produce plausible results that do not comply with the specified procedures, warranting dedicated study. We therefore identify tasks exhibiting EBU during benchmark construction and investigate two questions: how underuse on these tasks varies across models and harnesses, and whether revising reusable skill instructions can improve autonomous execution.

To study this problem systematically, we introduce MedSkill-E², a framework for evaluating and evolving skills for medical agents that comprises EBU-Bench and EBU-Evolve. EBU-Bench contains 331 tasks and their associated skills, combining tasks from public benchmarks with tasks authored around research questions. Evaluation is organized around reusable skills, each intended for a class of problems. We select task--skill pairs from this collection and evaluate nine advanced language models under two harnesses. Mean skill-use accuracy is 38.38\% when models use skills autonomously and 82.32\% with operational support; on EBU-Bench, all evaluated model--harness combinations exhibit EBU. Even with the model and skill held fixed, performance under autonomous use varies substantially across harnesses.

Operational support externally identifies a skill to use and provides general execution requirements for the task, but such guidance is usually unavailable in advance in practical settings. Notably, these supplementary instructions do not disclose any domain knowledge or operational steps specific to the task. We therefore propose EBU-Evolve to distill effective guidance from operational support into reusable skill instructions. The method revises a skill using paired execution feedback from the same task under autonomous use and operational support, and selects a revised version based on task performance under autonomous use. The revised skill can then be used directly for autonomous task solving without additional operational support for individual tasks. Experiments show that the revised skills improve overall skill-use accuracy under autonomous use on test tasks excluded from revision and selection, and continue to yield improvements when transferred across models or harnesses.

% generated by build_en_fable.py from ../en_fable_md/2_benchmark.md -- do not edit by hand
\section{EBU-Bench}

\subsection{Benchmark Construction}

EBU-Bench is a targeted challenge benchmark for skill underuse: it evaluates whether agents put available methods into practice. It combines tasks from public benchmarks with tasks authored around research questions; the public-benchmark task--skill pairs pass through three screening stages.

\textbf{Stage 1: task collection and capability screening.} We examine 11 public benchmarks and evaluation resources covering clinical calculation, medical image understanding, literature processing, biological data analysis, and clinical system operations \citep{NEURIPS2024_99e81750,Wang_2026,pmlr-v267-fallahpour25a,mitchener2025bixbenchcomprehensivebenchmarkllmbased,Jiang_2025,zhu-etal-2026-medmcp,liu2026physicianbenchevaluatingllmagents,Nair_2026,fa2026bioagent,workman2026scbenchevaluatingaiagents,wang2026drugclawdrugauditprimarysourcegroundedagent}. Tasks with the necessary data and tools are solved without a skill, and those not fully solved form the candidate pool.

\textbf{Stage 2: skill pairing and effectiveness validation.} For each task we retrieve and screen candidate skills from public skill libraries and methodological resources, organizing public methodological material into reusable procedural documents where necessary. Each candidate is then provided on its own with a prompt to perform its task-relevant operations, and pairs that can be completed under Operationally Supported Skill Use are retained.

\textbf{Stage 3: screening for an autonomous-use gap.} With the skill description provided alongside the task, the agent decides on its own whether and how to use the skill. For the same task and skill version, we retain tasks that are not completed under autonomous use but are completed under Operationally Supported Skill Use.

% \begin{wraptable}{r}{0.46\linewidth}
% \vspace{-1em}
% \centering\footnotesize
% \caption{Task composition of EBU-Bench.}\label{tab:task-composition}
% \begin{adjustbox}{max width=\linewidth}
% \begin{tabular}{llr}
% \toprule
% Work type & Task group & Tasks \\
% \midrule
% Clinical calculation & HAS-BLED & 8 \\
%  & PSI & 2 \\
%  & Delta gap & 30 \\
%  & Albumin-corrected Delta gap & 38 \\
% Biomedical data analysis & Multi-table integration and association analysis & 18 \\
%  & Paired matrix analysis & 2 \\
%  & Cohort and variant aggregation & 6 \\
%  & Expression–feature association & 9 \\
%  & Enrichment review & 51 \\
% Literature screening & Abstract screening & 61 \\
%  & Clinical evidence screening & 53 \\
% Functional evidence assessment & Variant functional evidence & 53 \\
% \textbf{Total} & \textbf{12 task groups} & \textbf{331} \\
% \bottomrule
% \end{tabular}
% \end{adjustbox}
% \vspace{-1em}
% \end{wraptable}

\begin{table}[!htb]
    \centering
    \caption{Task composition of EBU-Bench.}
    \label{tab:task-composition}

    \small
    \renewcommand{\arraystretch}{1.08}
    \begin{tabularx}{\linewidth}{
        @{}l
        >{\raggedright\arraybackslash}X
        r@{}
    }
        \toprule
        Work type & Task group & Tasks \\
        \midrule

        Clinical calculation
        & HAS-BLED & 8 \\
        & PSI & 2 \\
        & Delta gap & 30 \\
        & Albumin-corrected Delta gap & 38 \\

        \addlinespace[3pt]
        Biomedical data analysis
        & Multi-table integration and association analysis & 18 \\
        & Paired matrix analysis & 2 \\
        & Cohort and variant aggregation & 6 \\
        & Expression--feature association & 9 \\
        & Enrichment review & 51 \\

        \addlinespace[3pt]
        Literature screening
        & Abstract screening & 61 \\
        & Clinical evidence screening & 53 \\

        \addlinespace[3pt]
        Functional evidence assessment
        & Variant functional evidence & 53 \\

        \midrule
        \textbf{Total}
        & \textbf{12 task groups}
        & \textbf{331} \\
        \bottomrule
    \end{tabularx}
\end{table}
After the three screening stages, the retained public-benchmark tasks come from MedCalc-Bench and BixBench. To broaden coverage, we additionally author and expand tasks for literature screening, variant functional evidence, and enrichment review from common research questions, relevant domain skills, and public research materials; these require applying the skill's method under the specific conditions of the problem, such as screening studies against inclusion criteria, judging the applicability of functional evidence, or choosing the background for enrichment analysis. Varying problem conditions and resources yields 218 task pairings. Together with the 113 public-benchmark tasks, the two pathways yield 331 tasks after expansion, deduplication, and version freezing (Table~\ref{tab:task-composition}), each with its inputs, a skill, and scoring rules for the final output; sources, construction, and screening protocols are detailed in Appendix A.

\subsection{Task Definition and Evaluation Metrics}

\textbf{Task inputs and objectives.} Each task includes a problem description, the required data, the associated skill, and the tool environment needed for completion; the agent performs the requested calculations, analyses, or judgments and submits a final answer. Under Natural Skill Use, the associated skill is provided through the harness's standard mechanism, and the agent decides on its own whether to open it and how to use it (Figure~\ref{fig:overview}d).

\textbf{Task scoring.} Each task has a fixed scoring program: numerical answers are checked against predefined tolerances or intervals, categorical and structured answers against their rules. An answer scores 1 if it passes and 0 otherwise, and the mean score over the evaluated tasks is the skill-use accuracy. Scoring depends only on the final result; the agent chooses its own procedure. We also record from the execution trace whether the agent opens the associated skill.

\textbf{The EBU gap.} Operationally Supported Skill Use uses the same task and skill as Natural Skill Use but explicitly requires the agent to read the skill and perform its task-relevant operations, i.e., the operational support introduced above. With \(P_{\mathrm{natural}}(m,h)\) and \(P_{\mathrm{supported}}(m,h)\) the skill-use accuracies of model \(m\) under harness \(h\) in the two conditions, we define \(\Delta_{\mathrm{EBU}}(m,h)=P_{\mathrm{supported}}(m,h)-P_{\mathrm{natural}}(m,h)\). Using Operationally Supported Skill Use as an empirical reference, the EBU gap measures the task gains not yet realized under Natural Skill Use for a given model and harness. We report both accuracies and retain any negative differences; across harnesses, the gaps of the same model are averaged with equal weight. Skill-opening records and cross-harness differences support further analysis but do not enter the gap.

\textbf{Organizing tasks around skills.} Each associated skill corresponds to multiple tasks that apply the same method to different inputs and problem conditions (Table~\ref{tab:task-composition}), so that a method is evaluated across multiple instances; for skills with enough tasks, the tasks can be split into training, validation, and test sets for EBU-Evolve (§3).

\textbf{EBU-Bench Lite.} Because running the full benchmark across models, harnesses, and conditions is costly, we use a 154-task subset that keeps all 12 task groups: all 113 public-benchmark tasks plus one representative task per source-question--skill combination among the 218 authored tasks (41 tasks), which reduces the repeated weight of variants of the same source question. Representatives are selected independently of the evaluation results reported here (rules in Appendix A.4). Unless otherwise stated, the benchmark experiments use EBU-Bench Lite.

% generated by build_en_fable.py from ../en_fable_md/3_method.md -- do not edit by hand
\section{EBU-Evolve}

Figures~\ref{fig:overview}a and \ref{fig:overview}b show that general encouragement to use skills (Prompted Skill Use) remains insufficient to fully realize the benefits of an available skill. We propose EBU-Evolve, which derives concrete, generalizable revisions from the differences between executions of the same task under Natural Skill Use and Operationally Supported Skill Use, and distills them into persistent skill instructions. This addresses EBU on the individual task and, at the same time, enables agents to use the revised skill effectively on similar tasks that exhibit the same form of underuse, thereby realizing its value under Natural Skill Use.

Figure~\ref{fig:overview}c presents the overall EBU-Evolve workflow. When evolving a skill, we divide its tasks into three sets: the training set provides evidence for revision, the validation set determines the skill version, and the test set evaluates the selected version on tasks that took part in neither this revision nor the selection. Specifically, we use paired executions to determine what to revise, jointly revise the description and the body, and select the version with the highest mean task score under Natural Skill Use. Evolution leaves the model parameters and the skill's accompanying reference files, code, and data unchanged; only the skill's instruction text is updated.

\textbf{Distilling skill instructions from paired executions.} A single failure is insufficient to determine the direction of revision, whereas a successful execution of the same task under Operationally Supported Skill Use provides a reference. We therefore collect, for the same training task, the execution traces and scores under both conditions, using the same skill version, model, and harness. Building on trace-based reflective optimization \citep{agrawal2026gepa}, we pass these paired executions, together with the original skill and its reference materials, to a reflection model, which judges whether the agent accessed the relevant rules during autonomous execution and whether it applied them under the current task conditions, and proposes revision rules accordingly. Only revision rules supported by the original skill are used to generate candidates (Appendix B.1). We call this process \textbf{intervention-to-instruction distillation}: the improvements brought by operational support are written persistently into the skill.

\textbf{Jointly revising the description and body.} Distilled instructions must be accessed and correctly applied to improve autonomous performance. With progressive loading, the always-visible description indicates applicability; the body and references are read after opening. Clarifying rules therefore also requires guiding access to them. One edit produces a coordinated description and body, combined with the original into body-only, joint, and description-only candidates. Each is reviewed separately and compared with the original on the same Natural Skill Use validation tasks. Only a strict improvement replaces the original (Appendices B.2--B.4).

% generated by build_en_fable.py from ../en_fable_md/4_setup.md -- do not edit by hand
\section{Experimental Setup}

\subsection{Models and Execution Environments}

We evaluate nine models: GPT-5.5, Claude Opus 4.8, Gemini 3.8 Flash, DeepSeek V4 Pro, GLM-5.2, Atria-Dawn-Preview (Atria), Kimi K2.7 Code, Qwen3.8 Max, and Qwen3.8 27B FP8, each under two harnesses, Claude Code and Codex, which differ in resource discovery and tool interaction (request identifiers and harness versions in Appendix C).

Benchmark evaluations are executed through BenchFlow: each run uses an independent container and session with the data, computational tools, and file-operation interfaces required by the task, and external retrieval and data downloads are disabled to prevent answer leakage. All nine models are evaluated on the same 154 tasks described in §2.2; results of three models on the full benchmark of 331 tasks are reported in Appendix D.2.

\subsection{Skill Provision and Evaluation Conditions}

We ask not only whether a skill is provided but how much additional prompting an agent needs before using it, and define four conditions per task (Figure~\ref{fig:overview}a). The EBU gap is the difference between \textbf{Natural Skill Use} and \textbf{Operationally Supported Skill Use} (§2.2); \textbf{No Skill} and \textbf{Prompted Skill Use} measure performance without the associated skill and the change induced by a general reminder.

Under the conditions that provide a skill, the skill's name and description enter the context through the harness's native mechanism, and the body and reference files are read on demand. Natural Skill Use adds no task-level prompting, while retaining
the harness's built-in resource-use instructions. Prompted Skill Use adds only general encouragement to use the skill, without specifying solution steps, and tests whether merely reminding the agent suffices to realize the skill's benefits. Operationally Supported Skill Use further specifies the reading and execution requirements, providing a reference for the gains that remain unrealized. The three conditions with a skill use the same resource version and differ only in the support appended to the task prompt.

Operationally Supported Skill Use reuses a prompt fixed for each task group: for tasks derived from public benchmarks it requires reading the associated skill in full and applying its steps to the current inputs, and for authored tasks it requires reading the entry file and the reference materials it points to. Neither supplies the answer to the current task; because the two differ in specificity, we also report results by task source (Appendix C.4).

\subsection{Execution Protocol and Result Aggregation}

For each model, harness, and condition, we retain one run record for each of the 154 tasks, yielding 11,088 retained run records; all skill-use accuracies use 154 as the denominator. The initial time limit is either 900 or 3,600 seconds, depending on the batch. Runs that time out under the 900-second limit are rerun with a time limit of up to 7,200 seconds, and the rerun replaces the original record. Runs that time out or fail to submit a valid answer, including answers in a format the scorer cannot parse, receive a score of 0. Appendix C provides the detailed configuration, rerun scope, and distribution of invalid runs.
%For each model, harness, and condition we keep the score of one run per task (11,088 runs; all accuracies use 154 as the denominator). The initial time limit is 900 or 3,600 seconds depending on the batch; runs that time out at 900 seconds are rerun with up to 7,200 seconds, and the rerun replaces the original record. Runs that time out or submit no valid answer, including answers in a format the scorer cannot parse, score 0 (configuration, rerun scope, and invalid-run distribution in Appendix C).

%For each model, harness, and condition, we retain the score of one run on each of the 154 tasks, 11,088 runs in total, and all skill-use accuracies use 154 as the denominator. The initial time limit of a run is either 900 or 3,600 seconds depending on the batch; runs that time out under the 900-second limit are rerun with a budget of up to 7,200 seconds, and the rerun replaces the original record. Runs that time out or fail to submit a valid answer, including answers whose format does not meet the scoring requirements, are scored 0. The detailed configuration, the scope of reruns, and the distribution of invalid runs are given in Appendix C.

\subsection{Evaluation Setup for EBU-Evolve}

EBU-Evolve requires multiple tasks under the same skill to provide revision feedback and test the revisions. Because the task--skill pairs retained from public benchmarks do not support this split, we expand research questions from the literature into four skill groups: Abstract screening, Clinical evidence screening, Variant functional evidence, and Enrichment review. Training, validation, and test sets follow §3, with disjoint task IDs (splits and execution settings in Appendix B.7).

We evolve skills separately from the execution feedback of GLM-5.2, GPT-5.5, and Atria under Claude Code, with Claude Opus 4.8 performing reflection, editing, and review; each validation task is repeated three times and the version is selected by mean task score. At test time, the selected version is compared with the original skill under Natural Skill Use without operational support. To test reuse across configurations, each source's frozen version is also given to the other two source models and to DeepSeek V4 Pro, and tested with the same source model under Codex; all gains are relative to the original skill in the target configuration, and overall accuracies are aggregated over tasks.

The method comparison evaluates, under a common GLM-5.2/Claude Code configuration, the artifacts of the prompt-optimization method GEPA, the skill-optimization method SkillOpt, and EBU-Evolve; the first two are run with standard feedback (Std) and with paired evidence from the same task (Paired), under the same budget and editing permissions.

Further experiments examine editing scope, description and body resources, component contributions, and optimization stability. Both generic trace-reflection rewriting (hereafter, generic rewriting) and EBU-Evolve revise reusable skill instructions.  Generic rewriting uses feedback from Natural Skill Use traces.  EBU-Evolve pairs these traces with Operationally Supported Skill Use traces from the same training tasks, prioritizes cases in which support improves the task score, and diagnoses which skill rules the agent missed or misapplied. Independent optimization runs are denoted by $R$, and executions per validation task within a run by $K$; component comparisons use $R=3$ and $K=3$. Appendix B.8 reports local regressions, source-cluster uncertainty, and experiment-specific settings.

% generated by build_en_fable.py from ../en_fable_md/5_results.md -- do not edit by hand
\section{Experimental Results}

\subsection{EBU-Bench Results}

We first compare the skill-use accuracies of the nine models under Claude Code and Codex across the four conditions (Figure~\ref{fig:main-results}; per-cell values in Appendix Table~\ref{tab:D1}). Averaged with equal weight over the 18 model--harness combinations, the accuracies under No Skill, Natural Skill Use, and Operationally Supported Skill Use are 19.37\%, 38.38\%, and 82.32\%. All 18 combinations exhibit a positive EBU gap, from 8.44 to 73.38 pp with a mean of 43.94 pp (95\% CI: {[}39.57, 48.09{]}, bootstrap over tasks). The largest gap is GLM-5.2/Claude Code (73.38 pp); as task screening used GLM-5.1 from the same family, excluding GLM-5.2 leaves a largest gap of 71.43 pp (Qwen3.8 27B FP8/Claude Code). Even the best Natural Skill Use combination, Gemini 3.8 Flash/Codex, rises from 68.18\% to 76.62\% under Operationally Supported Skill Use.

\begin{figure}[!htb]\centering
\includegraphics[width=\textwidth]{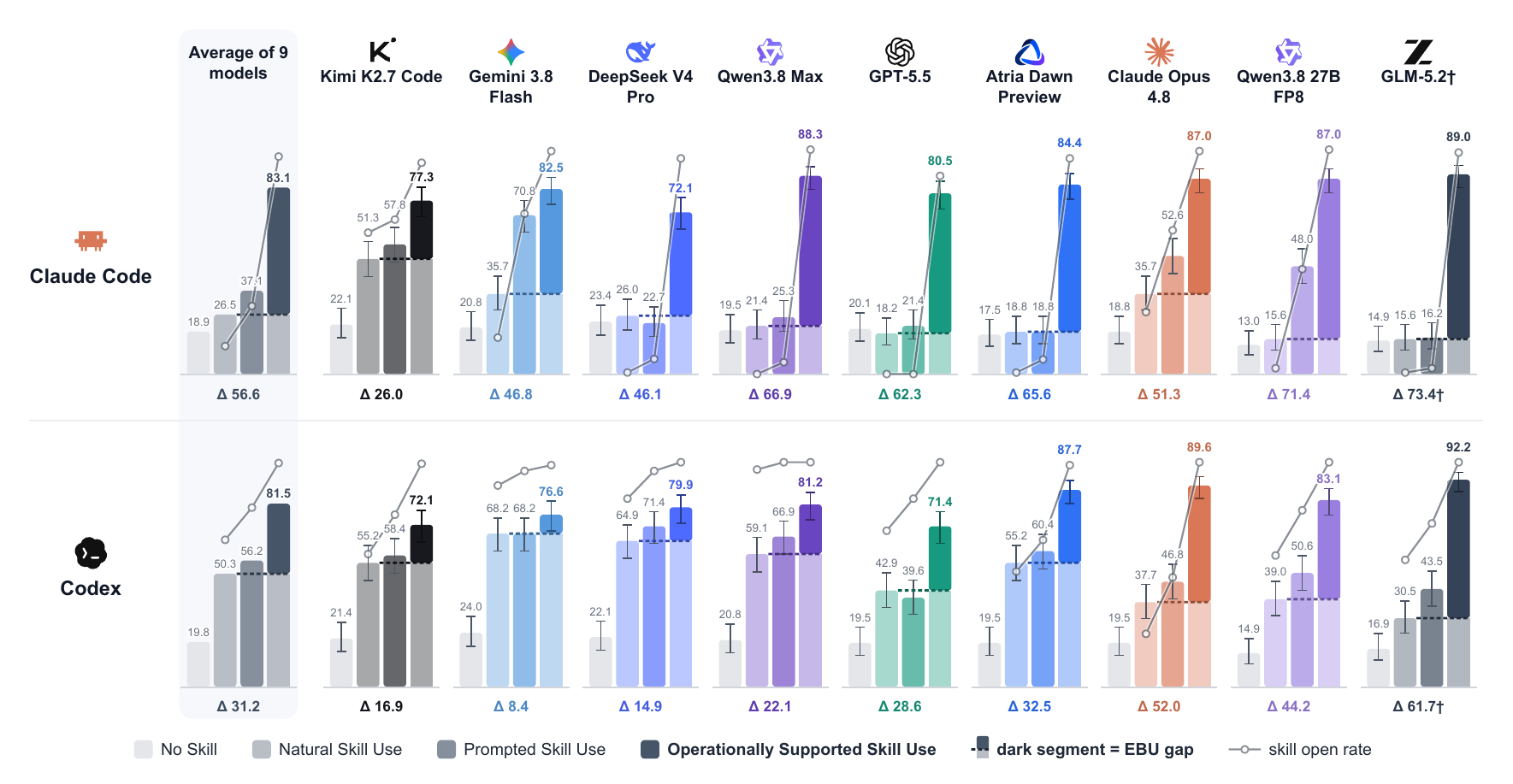}
\caption{Skill-use accuracy (\%) of nine models under two harnesses and four conditions. Models are ordered by increasing EBU gap averaged over the two harnesses; Δ below each group of bars is the EBU gap (pp). Error bars are Wilson 95\% intervals (n = 154). †: task screening used GLM-5.1 from the same model family.}\label{fig:main-results}
\end{figure}

\textbf{Autonomous skill access depends on the model--harness combination.} Cross-harness differences are concentrated in Natural Skill Use: the mean accuracies of Claude Code and Codex are 26.48\% and 50.29\%, a difference of 23.81 pp, whereas under No Skill and Operationally Supported Skill Use they differ by only 0.94 and 1.59 pp.~For the same model, Natural Skill Use accuracy differs by up to 38.96 pp across harnesses (DeepSeek V4 Pro), and the Spearman correlation between the harnesses' model rankings drops from 0.90 under No Skill to 0.51 under Natural Skill Use, consistent with the ranking changes Skill-Use observed for skill-use scores \citep{han2026skillusellmsactuallyuse}.

The same difference appears in resource access. A skill counts as opened once the target skill is read or loaded. Under Natural Skill Use the opening rates of Claude Code and Codex are 12.45\% and 65.42\%, whereas both approach saturation under Operationally Supported Skill Use (96.90\% and 99.64\%; Appendix E.1). Across the 18 combinations, opening rate and Natural Skill Use accuracy correlate at 0.91 (Pearson); Kimi K2.7 Code and Claude Opus 4.8, with similar opening rates under both harnesses, also show the smallest accuracy differences (3.90 and 1.95 pp). Thus, \textbf{autonomous skill utilization is a property of the model--harness combination; results and rankings under one harness do not transfer directly to another.}

\textbf{Opening a skill does not eliminate underuse.} Pairing runs by model, harness, and task, we decompose the net EBU gap by the change in opening status from Natural Skill Use to Operationally Supported Skill Use. Of the 56.64 pp gap of Claude Code, 54.69 pp comes from pairs that go from unopened to opened and only 1.95 pp from pairs opened under both conditions; for Codex the parts are 18.33 and 13.13 pp, the latter 42.0\% of its 31.24 pp gap. Yet on same-task pairs in which the skill is opened under both conditions, Operationally Supported Skill Use still raises accuracy from 69.23\% to 85.21\% for Claude Code and from 63.97\% to 84.15\% for Codex (counts in Appendix E.1), so opening records alone cannot tell whether a skill's benefit has been realized.

An Atria/Codex trace on a Delta gap task shows this directly: the model reads the skill's rule AG − 12, calls it ``unambiguous'', yet submits an answer computed from the memorized ΔAG − ΔHCO\(_3\); under Operationally Supported Skill Use the same model follows the resource and answers correctly (Appendix E.3). In medical tasks governed by a specification, a model may thus read the given rule and still substitute a familiar definition, producing a plausible result that does not comply with the provided specification.

\textbf{Prompted Skill Use closes only a small part of the EBU gap.} Prompted Skill Use encourages the model to use relevant resources without naming them. Relative to Natural Skill Use, accuracy rises from 38.38\% to 46.65\%, only 8.26 pp, closing 18.8\% of the mean EBU gap (by aggregate net successes) and leaving 35.68 pp to the 82.32\% of Operationally Supported Skill Use, which is higher in all 18 combinations. The traces also reveal a localized failure mode: under Prompted Skill Use, Qwen3.8 Max/Claude Code calls unregistered skill names in 60/154 runs and in 56 of them never opens the target skill yet submits an answer (Figure~\ref{fig:overview}a; Appendix E.2).

\subsection{EBU-Evolve Results}

\begin{figure}[!htbp]
  \centering
  \includegraphics[width=\textwidth]{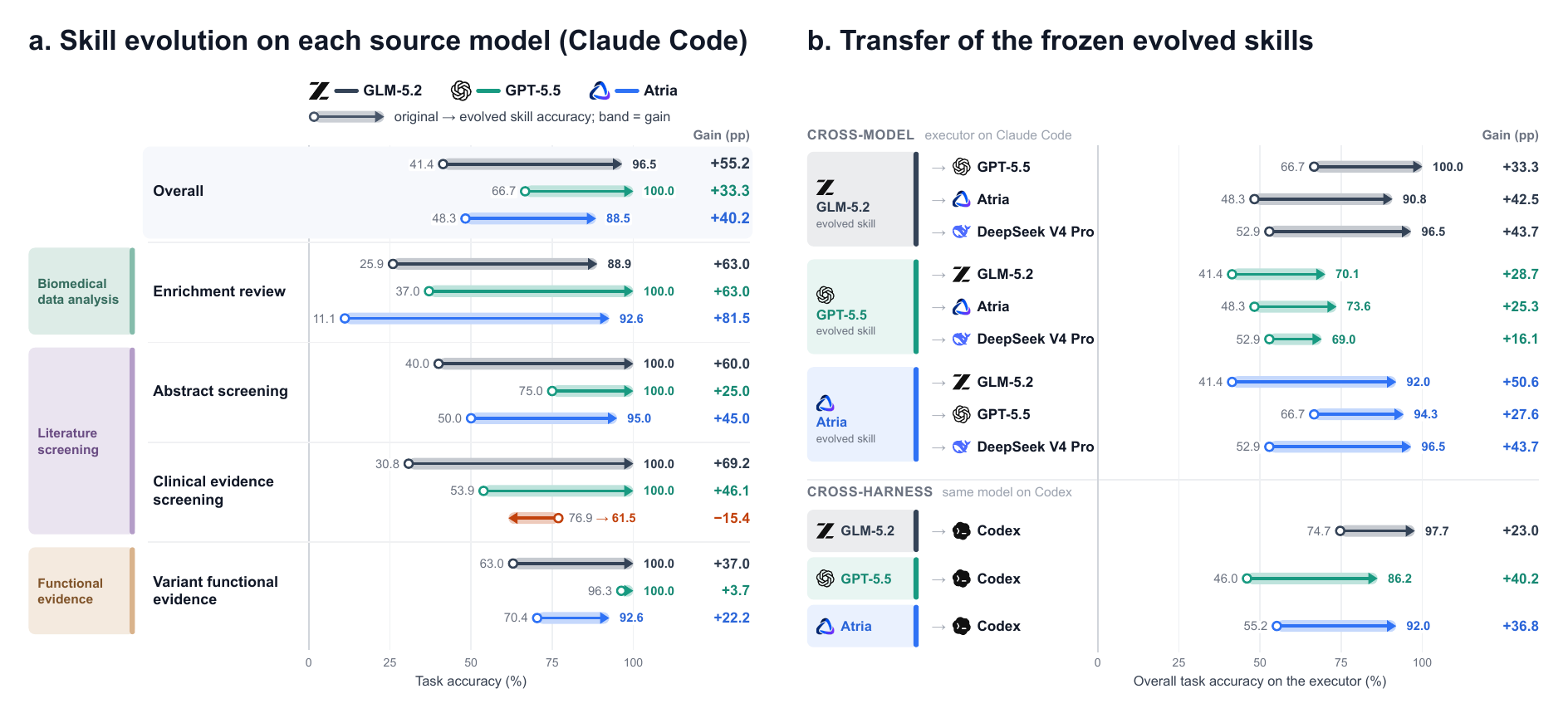}
  \caption{Skill evolution on source models and transfer of frozen skills. Orange marks a decrease in accuracy after evolution; in (b), the baseline of each target configuration is its own original skill.}
  \label{fig:ebu-evolve-main-v4}
\end{figure}

\textbf{Evolved skills improve Natural Skill Use performance and transfer to other models and harnesses.} EBU-Evolve raises Natural Skill Use accuracy on three source models spanning frontier open-weight, proprietary, and science-oriented models (Figure~\ref{fig:ebu-evolve-main-v4}a): on the 87 test tasks of the four skill groups (tasks of the same skill that took part in neither revision nor selection), skill-use accuracy improves by 55.17, 33.33, and 40.23 pp for GLM-5.2, GPT-5.5, and Atria. Distilling overlooked or ambiguously scoped rules into the skill thus improves Natural Skill Use on unseen tasks of the same skill.

The improvement transfers (Figure~\ref{fig:ebu-evolve-main-v4}b): the evolved skills can be frozen and used as-is elsewhere, with Natural Skill Use gains of 16.09--50.57 pp across models and 22.99--40.23 pp when the same model switches from Claude Code to Codex; all 12 transfer configurations gain, so one evolution can be reused across agents and harnesses without per-task guidance.

\textbf{Under the same budget and editing permissions, the skill revisions produced by EBU-Evolve are more effective than existing prompt-optimization and skill-optimization methods.} In the common GLM-5.2 test configuration, EBU-Evolve raises Natural Skill Use accuracy from 41.38\% to 96.55\%, a gain of 55.17 pp, against 2.30 and 18.39 pp for the best configurations of GEPA \citep{agrawal2026gepa} and SkillOpt \citep{yang2026skilloptexecutivestrategyselfevolving}, across all task types (Figure~\ref{fig:ebu-evolve-ablation-v7}a). Supplying the same paired feedback to the reflection modules of GEPA and SkillOpt (Paired) adds 2.30 and 10.34 pp over their standard settings (Std), so the Paired configuration also enriches the revision evidence of existing methods.

Two ablations test the EBU-Evolve design: decomposing the frozen GLM-5.2 artifact to separate the contributions of the description and body revisions, and comparing against generic trace-reflection rewriting to test the overall EBU-driven design. \textbf{Revising the description yields a larger gain with less added text, while jointly revising the description and body gives the highest skill-use accuracy} (Appendix B.8.1, Figure~\ref{fig:ebu-evolve-ablation-v7}b).

Description-only editing retains 85.42\% of the joint net gain. In separate controls, fixing the evolved description while removing body and reference access reduces accuracy from 91.95\% to 50.57\%; changing only the description with these resources fixed raises opening from 62.07\% to 100\%. EBU-Evolve also outperforms rule frontloading without execution feedback (Appendix B.8.1). Thus, description revision improves access, while body and reference resources support task completion.

\textbf{EBU-Evolve achieves higher accuracy than generic rewriting across independent optimizations.} In the initial comparison, EBU-Evolve attains 96.55\% accuracy, compared with 88.51\% for generic trace-reflection rewriting (Appendix B.8.2, Figure~\ref{fig:ebu-evolve-ablation-v7}c). To examine whether this advantage persists across independently optimized skills, we repeat reflection, editing, review, and validation selection three times. EBU-Evolve outperforms generic rewriting in all three runs, with a mean accuracy advantage of 10.73 pp. Abstract screening contributes the largest gain; across the remaining three task groups, the aggregate advantage is positive in every run and averages 5.47 pp (Appendices B.8.4--B.8.5). These results show that the advantage recurs across optimization runs, with its magnitude varying by task group.

\textbf{Paired evidence, gain-based case selection, and utilization diagnosis jointly improve revision quality.} Controlled ablations show that paired evidence raises mean accuracy; adding either selection or diagnosis improves it further, with the full method scoring highest (Appendix B.8.3). Thus, given same-task comparisons, selecting cases with score differences and identifying utilization problems helps produce effective revisions: selection focuses reflection on relevant cases, while diagnosis guides changes to access or rule application.

These results support the effectiveness of EBU-Evolve's design: distilling instructions through paired evidence, case selection, and utilization diagnosis improves autonomous skill utilization.

\textbf{Qualitative analysis reveals two EBU mechanisms.} Comparing traces of the same training task under the two conditions, one mechanism is failing to acquire the rule, e.g., answering after reading only the main SKILL.md without opening the reference file needed for the judgment; the other is recognizing the task condition but not carrying it into the computation, e.g., excluding assay-eligible but unannotated entities in Enrichment review (Appendix B.9, Figure~\ref{fig:distillation-case}).

\textbf{The evolved skills state the applicability conditions and execution order of existing rules more concretely, at a limited context cost of 227 added tokens per skill on average.} Rewriting also corrects the object of computation rather than merely inducing more reading: on a held-out Enrichment review test task, the original and evolved skills read the same files, yet only the evolved skill keeps all 12 eligible entities and answers correctly. As in §5.1, reading a skill does not guarantee following it; problems diagnosed in training are corrected on tasks outside the revision.

EBU-Evolve improves Atria-Dawn-Preview's overall Natural Skill Use accuracy while a few tasks regress (Figure~\ref{fig:ebu-evolve-main-v4}a). Three independent optimizations with Atria-Dawn all improve overall accuracy, with source-cluster analysis indicating uncertainty as task-source composition varies. Regressions involve both missing necessary materials and misapplying evidence-strength rules after reading, showing that aggregate gains can coexist with task-specific access and application errors (Appendix B.8.6).

% generated by build_en_fable.py from ../en_fable_md/6_related_work.md -- do not edit by hand
\section{Related Work}

\textbf{Agent evaluation and skill benchmarks.} Evaluation of LLMs and LLM agents covers interactive environments \citep{liu2024agentbench,merrill2026terminalbench} and specialized domains such as code repair, scientific data analysis, medical calculation, bioinformatics, and electronic health records \citep{jimenez2024swebench,chen2025scienceagentbench,NEURIPS2024_99e81750,mitchener2025bixbenchcomprehensivebenchmarkllmbased,Jiang_2025,fa2026bioagent}. Skill benchmarks instead evaluate the resources an agent is given: SkillsBench quantifies the benefit of a skill by paired evaluation of the same task with and without it and reports that this benefit varies across harnesses, a design we follow \citep{li2026skillsbenchbenchmarkingagentskills}; SWE-Skills-Bench uses the same comparison \citep{han2026sweskillsbenchagentskillsactually}, and SkillLearnBench tests whether generated skills transfer across instances of the same task \citep{zhong2026skilllearnbench}. Unlike these with/without designs, EBU-Bench holds the task and skill fixed and contrasts Natural Skill Use with Operationally Supported Skill Use, quantifying the part of a skill's benefit the agent does not realize on its own.

\textbf{From resource availability to effective use.} Providing information does not guarantee that a model uses it: performance depends on where the information sits in the context \citep{liu-etal-2024-lost}, and tool-use research separates whether a tool is invoked from how it is used \citep{ross-etal-2025-when2call}, while selective skill-invocation methods learn when to invoke a skill \citep{chen2026skillskiplearningselective}. For skills, controlled comparisons of loading, selection, and retrieval conditions show that explicit invocation instructions raise the reading rate while prescribed steps may still be skipped \citep{liu2026how,ying2026openskillevalautomaticallyauditingopen}; finer analyses decompose skill use into triggering, adherence, and boundaries, distinguish tested from satisfied constraints, or compare outcomes on tasks where the skill is actually retrieved \citep{han2026skillusellmsactuallyuse,tan2026skillcoveragetestadequacy,cho2026skillfollowingevaluatingactual}. The harness also affects performance and cost \citep{yao2026harnessbenchmeasuringharnesseffects}, and agents may follow repository context files without gaining accuracy \citep{gloaguen2026evaluatingagentsmdrepositorylevelcontext}. Access, adherence, and task-level benefit therefore need separate measurement; EBU-Bench measures underuse by the difference in skill-use accuracy rather than by reading or adherence, and examines its variation across harnesses with the model and skill fixed.

\textbf{Improving skills from execution feedback.} Retaining execution experience as an external resource is one route to continual improvement, from verbal reflections, cross-task rules, and programmatic skill libraries to prompt and context optimization with natural-language feedback \citep{NEURIPS2023_1b44b878,Zhao_2024,wang2024voyager,Yuksekgonul_2025,agrawal2026gepa,zhang2026agentic}; tool documentation and skill descriptions can likewise be rewritten from interaction problems or routing errors \citep{qu2025from,zhou2026singlerewritesufficesempirical}.For skills themselves, execution feedback has been used to discover new skills, summarize lessons, and select revised versions under validation feedback \citep{alzubi2026evoskill,ni2026trace2skilldistilltrajectorylocallessons,yang2026skilloptexecutivestrategyselfevolving,liu2026skillreviseimprovingllmauthoredagent}. These methods contrast different execution conditions: SkillGrad pairs current successes with failures under the initial skill \citep{wang2026skillgradoptimizingagentskills}; SkillGen derives procedures from successes and failures without a skill and verifies the net benefit of the newly generated resource \citep{ma2026skillgenverifiedinferencetimeagent}; and SkillAudit audits same-task traces with and without a skill to guide revision \citep{gao2026skillauditgroundtruthfreeskillevolution}. EBU-Evolve instead contrasts two use conditions of the same skill: successful executions under operational support locate the methods omitted or misapplied during autonomous use, and revisions are selected by Natural Skill Use outcomes.

% generated by build_en_fable.py from ../en_fable_md/7_conclusion.md -- do not edit by hand
\section{Conclusion}

We introduce MedSkill-E² to evaluate and improve medical agents' skill underuse. We define Enable But Underuse (EBU) as the phenomenon in which a skill supports task completion under operational support but its benefits are not fully realized autonomously. EBU-Bench combines public resources and tasks derived from real research questions through multistage screening, skill matching, and executable scoring. Its 331 challenging tasks span 12 groups across clinical calculation, biomedical data analysis, evidence-based literature screening, and functional evidence assessment. Comparing natural and supported use of the same tasks and skills quantifies unrealized benefits, providing a common basis for studying underuse. On the 154-task Lite benchmark, nine frontier models under two harnesses show a mean EBU gap of 43.94 pp. This gap varies across harnesses with the model and skill fixed; paired traces reveal failures in both instruction access and rule application. EBU-Evolve's intervention-to-instruction distillation incorporates guidance from these comparisons into descriptions and bodies, improving Natural Skill Use on held-out tasks of the same skill and transferring across models and harnesses. Component ablations and independent optimizations support this approach; mechanism comparisons show that descriptions and body resources complement each other in effective skill utilization.

In biomedicine, these findings distinguish knowledge availability from correct method execution. MedSkill-E² converts omitted conditions and misapplied rules into reusable, inspectable skill revisions, offering a path to improving medical evidence screening, assessment, and analysis. Expert review and validation in real medical workflows could test effectiveness across broader tasks, supporting more reliable execution in biomedical agent research and evidence processing.

MedSkill-E² focuses on biomedical tasks. Future work could extend EBU evaluation to other domains, investigate underuse across tasks and execution environments, and explore EBU-Evolve for continual skill self-evolution in general domains.

\clearpage
% Drop-in replacement for paper_手稿_latex/sections/statements.tex (ICLR 2027; does not count toward the page limit).
% Drafted 2026-09-26 from author-confirmed AI uses. Lines marked OPTIONAL assert facts the authors must confirm before keeping.

\subsection*{AI use statement}
In this work, we used generative AI tools to generate synthetic data, to clean, reformat, and de-identify datasets, to assist with implementing and running experiments, and to translate the manuscript from Chinese into English. We have not used generative AI tools to develop theoretical models or conceptual frameworks, formulate mathematical claims, provide ingredients for or write proofs, propose or refine hypotheses, design research methodology or experiments, support qualitative or thematic data analysis, or interpret results. Additionally, we used generative AI tools to write and edit software code, to summarize existing literature, to search for information, to identify and verify relevant literature and format references, to draft and revise parts of the related-work section, and to edit the paper to improve readability. We have reviewed all AI-assisted work. Every cited reference was checked against its official publication record and cross-checked against at least one independent bibliographic source, and each claim attributed to a cited work was verified against the original text. We take responsibility for the final content of this work, including text, claims, or artifacts produced with the aid of generative AI. Large language models are also the subject of this study and serve as components of EBU-Evolve (the reflection and editing models); these uses are part of the method and are described in Sections~3 and~4.

\subsection*{Ethics statement}
This work evaluates LLM agents on tasks derived from publicly released benchmarks and evaluation resources, together with tasks authored by the authors around research questions; each external resource was used under its license and access terms. Some of these research questions were drawn from research workflows within the authors' group; they were de-identified and manually reviewed before use, and the benchmark contains no identifiable personal information. Several source benchmarks involve clinical calculations or electronic health record environments; performance on EBU-Bench is not evidence that an agent or skill is safe or effective for clinical use, and the skills studied here should not be deployed in patient care without independent clinical validation.

\subsection*{Reproducibility statement}
Section~2 and Appendix~A describe how EBU-Bench tasks and their associated skills are constructed and how final outputs are scored. Section~4 and Appendix~C specify the evaluation configuration and run protocol, including model request identifiers and harness versions. Appendix~D.1 reports per-cell results for the main evaluation, Appendix~D.2 reports results on the full benchmark, and Appendix~E.1 gives the counts underlying the skill-opening and paired analyses. Section~3 and Appendix~B detail EBU-Evolve, including the data splits used for revision, selection, and testing.
% OPTIONAL (keep only if true): Anonymized code, task files, and evolved skills are provided in the supplementary material.

\bibliographystyle{iclr2027_conference}
\bibliography{refs_candidate}  % copy of section6/citation_audit/bib_script/refs_candidate.bib

@inproceedings{
liu2024agentbench,
title={AgentBench: Evaluating {LLM}s as Agents},
author={Xiao Liu and Hao Yu and Hanchen Zhang and Yifan Xu and Xuanyu Lei and Hanyu Lai and Yu Gu and Hangliang Ding and Kaiwen Men and Kejuan Yang and Shudan Zhang and Xiang Deng and Aohan Zeng and Zhengxiao Du and Chenhui Zhang and Sheng Shen and Tianjun Zhang and Yu Su and Huan Sun and Minlie Huang and Yuxiao Dong and Jie Tang},
booktitle={The Twelfth International Conference on Learning Representations},
year={2024},
url={https://openreview.net/forum?id=zAdUB0aCTQ}
}

@inproceedings{
merrill2026terminalbench,
title={Terminal-Bench: Benchmarking Agents on Hard, Realistic Tasks in Command Line Interfaces},
author={Mike A Merrill and Alexander Glenn Shaw and Nicholas Carlini and Boxuan Li and Harsh Raj and Ivan Bercovich and Lin Shi and Jeong Yeon Shin and Thomas Walshe and E. Kelly Buchanan and Junhong Shen and Guanghao Ye and Haowei Lin and Jason Poulos and Maoyu Wang and Marianna Nezhurina and Di Lu and Orfeas Menis Mastromichalakis and Zhiwei Xu and Zizhao Chen and Yue Liu and Robert Zhang and Leon Liangyu Chen and Anurag Kashyap and Jan-Lucas Uslu and Jeffrey Li and Jianbo Wu and Minghao Yan and Song Bian and Vedang Sharma and Ke Sun and Steven Dillmann and Akshay Anand and Andrew Lanpouthakoun and Bardia Koopah and Changran Hu and Etash Kumar Guha and Gabriel H. S. Dreiman and Jiacheng Zhu and Karl Krauth and Li Zhong and Niklas Muennighoff and Robert Kwesi Amanfu and Shangyin Tan and Shreyas Pimpalgaonkar and Tushar Aggarwal and Xiangning Lin and Xin Lan and Xuandong Zhao and Yiqing Liang and Yuanli Wang and Zilong Wang and Changzhi Zhou and David Heineman and Hange Liu and Harsh Trivedi and John Yang and Junhong Lin and Manish Shetty and Michael Yang and Nabil Omi and Negin Raoof and Shanda Li and Terry Yue Zhuo and Wuwei Lin and Yiwei Dai and Yuxin Wang and Wenhao Chai and Shang Zhou and Dariush Wahdany and Ziyu She and Jiaming Hu and Zhikang Dong and Yuxuan Zhu and Sasha Cui and Ahson Saiyed and Arinbj{\"o}rn Kolbeinsson and Jesse Hu and Christopher Michael Rytting and Ryan Marten and Yixin Wang and Jenia Jitsev and Alex Dimakis and Andy Konwinski and Ludwig Schmidt},
booktitle={The Fourteenth International Conference on Learning Representations},
year={2026},
url={https://openreview.net/forum?id=a7Qa4CcHak}
}

@inproceedings{
jimenez2024swebench,
title={{SWE}-bench: Can Language Models Resolve Real-world Github Issues?},
author={Carlos E Jimenez and John Yang and Alexander Wettig and Shunyu Yao and Kexin Pei and Ofir Press and Karthik R Narasimhan},
booktitle={The Twelfth International Conference on Learning Representations},
year={2024},
url={https://openreview.net/forum?id=VTF8yNQM66}
}

@inproceedings{
chen2025scienceagentbench,
title={ScienceAgentBench: Toward Rigorous Assessment of Language Agents for Data-Driven Scientific Discovery},
author={Ziru Chen and Shijie Chen and Yuting Ning and Qianheng Zhang and Boshi Wang and Botao Yu and Yifei Li and Zeyi Liao and Chen Wei and Zitong Lu and Vishal Dey and Mingyi Xue and Frazier N. Baker and Benjamin Burns and Daniel Adu-Ampratwum and Xuhui Huang and Xia Ning and Song Gao and Yu Su and Huan Sun},
booktitle={The Thirteenth International Conference on Learning Representations},
year={2025},
url={https://openreview.net/forum?id=6z4YKr0GK6}
}

@inproceedings{NEURIPS2024_99e81750,
 author = {Khandekar, Nikhil and Jin, Qiao and Xiong, Guangzhi and Dunn, Soren and Applebaum, Serina S and Anwar, Zain and Sarfo-Gyamfi, Maame and Safranek, Conrad W and Anwar, Abid A and Zhang, Andrew and Gilson, Aidan and Singer, Maxwell B and Dave, Amisha and Taylor, Andrew and Zhang, Aidong and Chen, Qingyu and Lu, Zhiyong},
 booktitle = {Advances in Neural Information Processing Systems},
 doi = {10.52202/079017-2690},
 editor = {A. Globerson and L. Mackey and D. Belgrave and A. Fan and U. Paquet and J. Tomczak and C. Zhang},
 pages = {84730--84745},
 publisher = {Curran Associates, Inc.},
 title = {MedCalc-Bench: Evaluating Large Language Models for Medical Calculations},
 url = {https://proceedings.neurips.cc/paper_files/paper/2024/file/99e81750f3fdfcaf9613db2dbf4bd623-Paper-Datasets_and_Benchmarks_Track.pdf},
 volume = {37},
 year = {2024}
}

@misc{mitchener2025bixbenchcomprehensivebenchmarkllmbased,
      title={BixBench: a Comprehensive Benchmark for LLM-based Agents in Computational Biology}, 
      author={Ludovico Mitchener and Jon M Laurent and Alex Andonian and Benjamin Tenmann and Siddharth Narayanan and Geemi P Wellawatte and Andrew White and Lorenzo Sani and Samuel G Rodriques},
      year={2025},
      eprint={2503.00096},
      archivePrefix={arXiv},
      primaryClass={q-bio.QM},
      url={https://arxiv.org/abs/2503.00096}, 
}

@article{Jiang_2025, title={MedAgentBench: A Virtual EHR Environment to Benchmark Medical LLM Agents}, volume={2}, ISSN={2836-9386}, url={http://dx.doi.org/10.1056/AIdbp2500144}, DOI={10.1056/aidbp2500144}, number={9}, journal={NEJM AI}, publisher={Massachusetts Medical Society}, author={Jiang, Yixing and Black, Kameron C. and Geng, Gloria and Park, Danny and Zou, James and Ng, Andrew Y. and Chen, Jonathan H.}, year={2025}, month = aug }

@inproceedings{
fa2026bioagent,
title={BioAgent Bench: An {AI} Agent Evaluation Suite for Bioinformatics},
author={Dionizije Fa and Marko {\v{C}}uljak and Bruno Pand{\v{z}}a and Mateo {\v{C}}upi{\'c}},
booktitle={Forty-third International Conference on Machine Learning},
year={2026},
url={https://openreview.net/forum?id=2iFauBXg7Y}
}

@misc{li2026skillsbenchbenchmarkingagentskills,
      title={SkillsBench: Benchmarking How Well Agent Skills Work Across Diverse Tasks}, 
      author={Xiangyi Li and Yimin Liu and Wenbo Chen and Bingran You and Zonglin Di and Yifeng He and Shenghan Zheng and Kyoung Whan Choe and Jiankai Sun and Shuyi Wang and Chujun Tao and Binxu Li and Xuandong Zhao and Hejia Geng and Xiaojun Wu and Junwei Zhou and Xiaokun Chen and Hanwen Xing and Yubo Li and Qunhong Zeng and Di Wang and Yuanli Wang and Roey Ben Chaim and Penghao Jiang and Haotian Shen and Luyang Kong and Xinyi Liu and Runhui Wang and Xuanqing Liu and Jiachen Li and Xin Lan and Yueqian Lin and Wengao Ye and Junwei He and Songlin Li and Yue Zhang and Yipeng Gao and Yijiang Li and Ze Ma and Liqiang Jing and Tianyu Wang and Kaixin Li and Yiqi Xue and Haoran Lyu and Yizhuo He and Yuchen Tian and Shutong Wu and Bowei Wang and Yixuan Gao and Bo Chen and Litong Liu and Sikai Cheng and Jiajun Bao and Shuaicheng Tong and Shuwen Xu and Terry Yue Zhuo and Tinghan Ye and Qi Qi and Miao Li and Longtai Liao and Zelin Tan and Chang Shi and Xilin Tang and Srinath Tankasala and Boqin Yuan and Yaoyao Qian and Jianhong Tu and Chenguang Wang and Yizhou Sun and Wei Wang and Aaron Taylor and Ziyue Yang and Changkun Guan and Zhikang Dong and Xinyu Zhang and Steven Dillmann and Han-chung Lee and Dawn Song},
      year={2026},
      eprint={2602.12670},
      archivePrefix={arXiv},
      primaryClass={cs.AI},
      url={https://arxiv.org/abs/2602.12670}, 
}

@misc{han2026sweskillsbenchagentskillsactually,
      title={SWE-Skills-Bench: Do Agent Skills Actually Help in Real-World Software Engineering?}, 
      author={Tingxu Han and Yi Zhang and Wei Song and Chunrong Fang and Zhenyu Chen and Youcheng Sun and Lijie Hu},
      year={2026},
      eprint={2603.15401},
      archivePrefix={arXiv},
      primaryClass={cs.SE},
      url={https://arxiv.org/abs/2603.15401}, 
}

@inproceedings{
zhong2026skilllearnbench,
title={SkillLearnBench: Benchmarking Continual Learning Methods for Agent Skill Generation on Real-World Tasks},
author={Shanshan Zhong and Yi Lu and Jingjie Ning and Yibing Wan and Lihan Feng and Yuyi Ao and Leonardo F. R. Ribeiro and Markus Dreyer and Sean Ammirati and Chenyan Xiong},
booktitle={Third Conference on Language Modeling},
year={2026},
url={https://openreview.net/forum?id=ulrZEJrNXA}
}

@article{liu-etal-2024-lost,
    title = "Lost in the Middle: How Language Models Use Long Contexts",
    author = "Liu, Nelson F.  and
      Lin, Kevin  and
      Hewitt, John  and
      Paranjape, Ashwin  and
      Bevilacqua, Michele  and
      Petroni, Fabio  and
      Liang, Percy",
    journal = "Transactions of the Association for Computational Linguistics",
    volume = "12",
    year = "2024",
    address = "Cambridge, MA",
    publisher = "MIT Press",
    url = "https://aclanthology.org/2024.tacl-1.9/",
    doi = "10.1162/tacl_a_00638",
    pages = "157--173"
}

@inproceedings{ross-etal-2025-when2call,
    title = "{W}hen2{C}all: When (not) to Call Tools",
    author = "Ross, Hayley  and
      Mahabaleshwarkar, Ameya Sunil  and
      Suhara, Yoshi",
    editor = "Chiruzzo, Luis  and
      Ritter, Alan  and
      Wang, Lu",
    booktitle = "Proceedings of the 2025 Conference of the Nations of the Americas Chapter of the Association for Computational Linguistics: Human Language Technologies (Volume 1: Long Papers)",
    month = apr,
    year = "2025",
    address = "Albuquerque, New Mexico",
    publisher = "Association for Computational Linguistics",
    url = "https://aclanthology.org/2025.naacl-long.174/",
    doi = "10.18653/v1/2025.naacl-long.174",
    pages = "3391--3409",
    ISBN = "979-8-89176-189-6"
}

@misc{chen2026skillskiplearningselective,
      title={Skill or Skip? Learning Selective Skill Invocation in Agentic Tasks via Dual-Granularity Preference Learning}, 
      author={Chishui Chen and Jiaye Lin and Te Sun and Yi Yang and Junxi Wang and Cong Qin and Yangen Hu and Lu Pan and Ke Zeng},
      year={2026},
      eprint={2606.00510},
      archivePrefix={arXiv},
      primaryClass={cs.CL},
      url={https://arxiv.org/abs/2606.00510}, 
}

@inproceedings{
liu2026how,
title={How Well Do Agentic Skills Work in the Wild: Benchmarking {LLM} Skill Usage in Realistic Settings},
author={Yujian Liu and Jiabao Ji and Li An and Tommi Jaakkola and Yang Zhang and Shiyu Chang},
booktitle={Third Conference on Language Modeling},
year={2026},
url={https://openreview.net/forum?id=axGWHkyD1z}
}

@misc{ying2026openskillevalautomaticallyauditingopen,
      title={OpenSkillEval: Automatically Auditing the Open Skill Ecosystem for LLM Agents}, 
      author={Jiahao Ying and Boxian Ai and Wei Tang and Siyuan Liu and Yixin Cao},
      year={2026},
      eprint={2605.23657},
      archivePrefix={arXiv},
      primaryClass={cs.CL},
      url={https://arxiv.org/abs/2605.23657}, 
}

@misc{han2026skillusellmsactuallyuse,
      title={Skill-Use: Can LLMs Actually Use Skills in Agentic Harnesses?}, 
      author={Jinyi Han and Yuanjian Xu and Ying Liao and Xinyi Wang and Zishang Jiang and Zixiang Di and Fanyang Lu and Zhichao Hu and Yanghua Xiao},
      year={2026},
      eprint={2608.04828},
      archivePrefix={arXiv},
      primaryClass={cs.CL},
      url={https://arxiv.org/abs/2608.04828}, 
}

@misc{tan2026skillcoveragetestadequacy,
      title={Skill Coverage: A Test Adequacy Metric for Agent Skills}, 
      author={Boyin Tan and Xiaowei Huang and Youcheng Sun},
      year={2026},
      eprint={2606.20659},
      archivePrefix={arXiv},
      primaryClass={cs.AI},
      url={https://arxiv.org/abs/2606.20659}, 
}

@misc{cho2026skillfollowingevaluatingactual,
      title={Skill Following: Evaluating Actual Skill Use in Retrieval-Enabled LLM Agents}, 
      author={Seonghyeon Cho and Chanjun Park},
      year={2026},
      eprint={2609.00549},
      archivePrefix={arXiv},
      primaryClass={cs.CL},
      url={https://arxiv.org/abs/2609.00549}, 
}

@misc{yao2026harnessbenchmeasuringharnesseffects,
      title={Harness-Bench: Measuring Harness Effects across Models in Realistic Agent Workflows}, 
      author={Yilun Yao and Xinyu Tan and Chao-Hsuan Liu and Yaoming Li and Zhengyang Wang and Wenhan Yu and Zhewen Tan and Yuxuan Tian and Guangxiang Zhao and Lin Sun and Xiangzheng Zhang and Tong Yang},
      year={2026},
      eprint={2605.27922},
      archivePrefix={arXiv},
      primaryClass={cs.AI},
      url={https://arxiv.org/abs/2605.27922}, 
}

@misc{gloaguen2026evaluatingagentsmdrepositorylevelcontext,
      title={Evaluating AGENTS.md: Are Repository-Level Context Files Helpful for Coding Agents?}, 
      author={Thibaud Gloaguen and Niels Mündler and Mark Müller and Veselin Raychev and Martin Vechev},
      year={2026},
      eprint={2602.11988},
      archivePrefix={arXiv},
      primaryClass={cs.SE},
      url={https://arxiv.org/abs/2602.11988}, 
}

@inproceedings{NEURIPS2023_1b44b878,
 author = {Shinn, Noah and Cassano, Federico and Gopinath, Ashwin and Narasimhan, Karthik and Yao, Shunyu},
 booktitle = {Advances in Neural Information Processing Systems},
 doi = {10.52202/075280-0377},
 editor = {A. Oh and T. Naumann and A. Globerson and K. Saenko and M. Hardt and S. Levine},
 pages = {8634--8652},
 publisher = {Curran Associates, Inc.},
 title = {Reflexion: language agents with verbal reinforcement learning},
 url = {https://proceedings.neurips.cc/paper_files/paper/2023/file/1b44b878bb782e6954cd888628510e90-Paper-Conference.pdf},
 volume = {36},
 year = {2023}
}

@article{Zhao_2024, title={ExpeL: LLM Agents Are Experiential Learners}, volume={38}, ISSN={2159-5399}, url={http://dx.doi.org/10.1609/aaai.v38i17.29936}, DOI={10.1609/aaai.v38i17.29936}, number={17}, journal={Proceedings of the AAAI Conference on Artificial Intelligence}, publisher={Association for the Advancement of Artificial Intelligence (AAAI)}, author={Zhao, Andrew and Huang, Daniel and Xu, Quentin and Lin, Matthieu and Liu, Yong-Jin and Huang, Gao}, year={2024}, month = mar, pages={19632–19642} }

@article{
wang2024voyager,
title={Voyager: An Open-Ended Embodied Agent with Large Language Models},
author={Guanzhi Wang and Yuqi Xie and Yunfan Jiang and Ajay Mandlekar and Chaowei Xiao and Yuke Zhu and Linxi Fan and Anima Anandkumar},
journal={Transactions on Machine Learning Research},
issn={2835-8856},
year={2024},
url={https://openreview.net/forum?id=ehfRiF0R3a},
note={}
}

@article{Yuksekgonul_2025, title={Optimizing generative AI by backpropagating language model feedback}, volume={639}, ISSN={1476-4687}, url={http://dx.doi.org/10.1038/s41586-025-08661-4}, DOI={10.1038/s41586-025-08661-4}, number={8055}, journal={Nature}, publisher={Springer Science and Business Media LLC}, author={Yuksekgonul, Mert and Bianchi, Federico and Boen, Joseph and Liu, Sheng and Lu, Pan and Huang, Zhi and Guestrin, Carlos and Zou, James}, year={2025}, month = mar, pages={609–616} }

@inproceedings{
agrawal2026gepa,
title={{GEPA}: Reflective Prompt Evolution Can Outperform Reinforcement Learning},
author={Lakshya A Agrawal and Shangyin Tan and Dilara Soylu and Noah Ziems and Rishi Khare and Krista Opsahl-Ong and Arnav Singhvi and Herumb Shandilya and Michael J Ryan and Meng Jiang and Christopher Potts and Koushik Sen and Alex Dimakis and Ion Stoica and Dan Klein and Matei Zaharia and Omar Khattab},
booktitle={The Fourteenth International Conference on Learning Representations},
year={2026},
url={https://openreview.net/forum?id=RQm2KQTM5r}
}

@inproceedings{
zhang2026agentic,
title={Agentic Context Engineering: Evolving Contexts for Self-Improving Language Models},
author={Qizheng Zhang and Changran Hu and Shubhangi Upasani and Boyuan Ma and Fenglu Hong and Vamsidhar Kamanuru and Jay Rainton and Chen Wu and Mengmeng Ji and Hanchen Li and Urmish Thakker and James Zou and Kunle Olukotun},
booktitle={The Fourteenth International Conference on Learning Representations},
year={2026},
url={https://openreview.net/forum?id=eC4ygDs02R}
}

@inproceedings{
qu2025from,
title={From Exploration to Mastery: Enabling {LLM}s to Master Tools via Self-Driven Interactions},
author={Changle Qu and Sunhao Dai and Xiaochi Wei and Hengyi Cai and Shuaiqiang Wang and Dawei Yin and Jun Xu and Ji-Rong Wen},
booktitle={The Thirteenth International Conference on Learning Representations},
year={2025},
url={https://openreview.net/forum?id=QKBu1BOAwd}
}

@misc{zhou2026singlerewritesufficesempirical,
      title={A Single Rewrite Suffices: Empirical Lessons from Production Skill Description Optimization}, 
      author={Yangqiaoyu Zhou and Mohammad Alqudah and Kwei-Herng Lai and Aaron Halfaker and Yingqi Xiong and Yaar Harari},
      year={2026},
      eprint={2606.30775},
      archivePrefix={arXiv},
      primaryClass={cs.CL},
      url={https://arxiv.org/abs/2606.30775}, 
}

@inproceedings{
alzubi2026evoskill,
title={EvoSkill: Automated Skill Discovery for Multi-Agent Systems},
author={Salaheddin Alzubi and Noah Provenzano and Jaydon Bingham and Christian Alexander Calvo and Weiyuan Chen and Tu Vu},
booktitle={Third Conference on Language Modeling},
year={2026},
url={https://openreview.net/forum?id=tLvlDr1GWD}
}

@misc{ni2026trace2skilldistilltrajectorylocallessons,
      title={Trace2Skill: Distill Trajectory-Local Lessons into Transferable Agent Skills}, 
      author={Jingwei Ni and Yihao Liu and Xinpeng Liu and Yutao Sun and Mengyu Zhou and Pengyu Cheng and Dexin Wang and Erchao Zhao and Xiaoxi Jiang and Guanjun Jiang},
      year={2026},
      eprint={2603.25158},
      archivePrefix={arXiv},
      primaryClass={cs.AI},
      url={https://arxiv.org/abs/2603.25158}, 
}

@misc{yang2026skilloptexecutivestrategyselfevolving,
      title={SkillOpt: Executive Strategy for Self-Evolving Agent Skills}, 
      author={Yifan Yang and Ziyang Gong and Weiquan Huang and Qihao Yang and Ziwei Zhou and Zisu Huang and Yan Li and Xuemei Gao and Qi Dai and Bei Liu and Kai Qiu and Yuqing Yang and Dongdong Chen and Xue Yang and Chong Luo},
      year={2026},
      eprint={2605.23904},
      archivePrefix={arXiv},
      primaryClass={cs.AI},
      url={https://arxiv.org/abs/2605.23904}, 
}

@misc{liu2026skillreviseimprovingllmauthoredagent,
      title={SkillRevise: Improving LLM-Authored Agent Skills via Trace-Conditioned Skill Revision}, 
      author={Yuxuan Liu and Zhaochen Su and Lingyun Xie and Yuhao Zhang and Qing Zong and Jiahe Guo and Zhongwei Xie and Yiyan Ji and Yauwai Yim and Hongyu Luo and Xiyu Ren and Ruan, Chenyu and Haoran Li and Yangqiu Song},
      year={2026},
      eprint={2606.01139},
      archivePrefix={arXiv},
      primaryClass={cs.AI},
      url={https://arxiv.org/abs/2606.01139}, 
}

@misc{wang2026skillgradoptimizingagentskills,
      title={SkillGrad: Optimizing Agent Skills Like Gradient Descent}, 
      author={Hanyu Wang and Yifan Lan and Bochuan Cao and Lu Lin and Jinghui Chen},
      year={2026},
      eprint={2605.27760},
      archivePrefix={arXiv},
      primaryClass={cs.AI},
      url={https://arxiv.org/abs/2605.27760}, 
}

@misc{ma2026skillgenverifiedinferencetimeagent,
      title={SkillGen: Verified Inference-Time Agent Skill Synthesis}, 
      author={Yuchen Ma and Yue Huang and Han Bao and Haomin Zhuang and Swadheen Shukla and Michel Galley and Xiangliang Zhang and Stefan Feuerriegel},
      year={2026},
      eprint={2605.10999},
      archivePrefix={arXiv},
      primaryClass={cs.LG},
      url={https://arxiv.org/abs/2605.10999}, 
}

@misc{gao2026skillauditgroundtruthfreeskillevolution,
      title={SkillAudit: Ground-Truth-Free Skill Evolution via Paired Trajectory Auditing}, 
      author={Haowen Gao and Haoran Chen and Can Wang and Shasha Guo and Liang Pang and Zhaoyang Liu and Huawei Shen and Xueqi Cheng},
      year={2026},
      eprint={2606.14239},
      archivePrefix={arXiv},
      primaryClass={cs.AI},
      url={https://arxiv.org/abs/2606.14239}, 
}

@misc{xu2026empiricalstudyagentskills,
      title={An Empirical Study of Agent Skills for Healthcare: Practice, Gaps, and Governance}, 
      author={Gelei Xu and Ningzhi Tang and Xueyang Li and Toby Jia-Jun Li and Zhi Zheng and Wei Jin and Yiyu Shi},
      year={2026},
      eprint={2605.02709},
      archivePrefix={arXiv},
      primaryClass={cs.AI},
      url={https://arxiv.org/abs/2605.02709}, 
}

@article{Wang_2026, title={Benchmarking and developing large language models using one million clinical trials}, ISSN={2398-6352}, url={http://dx.doi.org/10.1038/s41746-026-02933-7}, DOI={10.1038/s41746-026-02933-7}, journal={npj Digital Medicine}, publisher={Springer Science and Business Media LLC}, author={Wang, Zifeng and Lin, Jiacheng and Jin, Qiao and Gao, Junyi and Pradeepkumar, Jathurshan and Jiang, Pengcheng and Lu, Zhiyong and Sun, Jimeng}, year={2026}, month = jul }

@InProceedings{pmlr-v267-fallahpour25a,
  title = 	 {{M}ed{RAX}: Medical Reasoning Agent for Chest X-ray},
  author =       {Fallahpour, Adibvafa and Ma, Jun and Munim, Alif and Lyu, Hongwei and Wang, Bo},
  booktitle = 	 {Proceedings of the 42nd International Conference on Machine Learning},
  pages = 	 {15661--15676},
  year = 	 {2025},
  editor = 	 {Singh, Aarti and Fazel, Maryam and Hsu, Daniel and Lacoste-Julien, Simon and Berkenkamp, Felix and Maharaj, Tegan and Wagstaff, Kiri and Zhu, Jerry},
  volume = 	 {267},
  series = 	 {Proceedings of Machine Learning Research},
  month = 	 {13--19 Jul},
  publisher =    {PMLR},
  url = 	 {https://proceedings.mlr.press/v267/fallahpour25a.html}
}

@inproceedings{zhu-etal-2026-medmcp,
    title = "{M}ed{MCP}-Calc: Benchmarking {LLM}s for Realistic Medical Calculator Scenarios via {MCP} Integration",
    author = "Zhu, Yakun  and
      Huang, Yutong  and
      Qin, Shengqian  and
      Huang, Zhongzhen  and
      Zhang, Shaoting  and
      Zhang, Xiaofan",
    editor = "Liakata, Maria  and
      Moreira, Viviane P.  and
      Zhang, Jiajun  and
      Jurgens, David",
    booktitle = "Proceedings of the 64th Annual Meeting of the {A}ssociation for {C}omputational {L}inguistics (Volume 1: Long Papers)",
    month = jul,
    year = "2026",
    address = "San Diego, California, United States",
    publisher = "Association for Computational Linguistics",
    url = "https://aclanthology.org/2026.acl-long.221/",
    doi = "10.18653/v1/2026.acl-long.221",
    pages = "4845--4873",
    ISBN = "979-8-89176-390-6"
}

@misc{liu2026physicianbenchevaluatingllmagents,
      title={PhysicianBench: Evaluating LLM Agents in Real-World EHR Environments}, 
      author={Ruoqi Liu and Imran Q. Mohiuddin and Austin J. Schoeffler and Kavita Renduchintala and Ashwin Nayak and Prasantha L. Vemu and Shivam C. Vedak and Kameron C. Black and John L. Havlik and Isaac Ogunmola and Stephen P. Ma and Roopa Dhatt and Jonathan H. Chen},
      year={2026},
      eprint={2605.02240},
      archivePrefix={arXiv},
      primaryClass={cs.AI},
      url={https://arxiv.org/abs/2605.02240}, 
}

@article{Nair_2026, title={Agentic systems are adept at solving well-scoped, verifiable problems in computational biology}, url={http://dx.doi.org/10.64898/2026.04.06.716850}, DOI={10.64898/2026.04.06.716850}, publisher={openRxiv}, author={Nair, Surag and Gunsalus, Laura and Orcutt-Jahns, Brian and Rossen, Jordan and Lal, Avantika and De Donno, Carlo and {\c{C}}elik, Muhammed Hasan and Fletez-Brant, Kipper and Xie, Xiaoman and Corrada Bravo, Hector and Eraslan, Gokcen}, year={2026}, month = apr }

@misc{workman2026scbenchevaluatingaiagents,
      title={scBench: Evaluating AI Agents on Single-Cell RNA-seq Analysis}, 
      author={Kenny Workman and Zhen Yang and Harihara Muralidharan and Aidan Abdulali and Hannah Le},
      year={2026},
      eprint={2602.09063},
      archivePrefix={arXiv},
      primaryClass={q-bio.GN},
      url={https://arxiv.org/abs/2602.09063}, 
}

@misc{wang2026drugclawdrugauditprimarysourcegroundedagent,
      title={DrugClaw and DrugAudit: A Primary-Source-Grounded Agent and Authority-Aware Benchmark for Drug-Information Question Answering}, 
      author={Qing Wang and Bo Li and Jialu Liang and Daling Shi and Bob Zhang and Qianqian Song},
      year={2026},
      eprint={2606.01434},
      archivePrefix={arXiv},
      primaryClass={cs.CL},
      url={https://arxiv.org/abs/2606.01434}, 
}

\appendix
\counterwithin{table}{section}  % appendix tables: A.1, A.2, ...
\counterwithin{figure}{section}
\clearpage
% generated by build_en_fable.py from ../en_fable_md/8_appendix_A.md -- do not edit by hand
\section{Construction and Scoring of EBU-Bench}

The 331 tasks of EBU-Bench come from two pathways: 113 tasks screened from public benchmarks (A.1) and 218 tasks authored and expanded around research questions (A.2). A.3 gives the scoring rules of each task group, and A.4 describes the selection of representative tasks for EBU-Bench Lite.

\subsection{Tasks Derived from Public Benchmarks}

\textbf{Sources.} We registered 11 public benchmarks and evaluation resources with 9,646 task entries in total (Table~\ref{tab:A1}). Some entries lack the data or tools needed to run them; the first round actually evaluated 9,305 records from 5 benchmarks.

\begin{table}[!htb]
    \centering
    \caption{Registered public benchmark sources.}
    \label{tab:A1}

    \small
    \begin{tabular*}{\linewidth}{@{\extracolsep{\fill}}lrr@{}}
        \toprule
        Source & Registered entries & Records evaluated in round 1 \\
        \midrule
        MedCalc-Bench   & 1,100 & 1,100 \\
        TrialPanorama   & 5,200 & 5,200 \\
        ChestAgentBench & 2,500 & 2,500 \\
        BixBench        &   205 &   205 \\
        MedAgentBench   &   300 &   300 \\
        MedMCP-Calc     &   118 &   --- \\
        PhysicianBench &   100 &   --- \\
        CompBioBench   &   100 &   --- \\
        BioAgentBench  &    10 &   --- \\
        scBench        &     6 &   --- \\
        DrugClaw       &     7 &   --- \\
        \midrule
        \textbf{Total} & \textbf{9,646} & \textbf{9,305} \\
        \bottomrule
    \end{tabular*}
\end{table}
\textbf{Stage 1: No Skill screening.} This stage has two rounds. The first round evaluates all runnable entries without providing a skill, using DeepSeek V4 Pro (GPT-4o for chest X-ray tasks), and records 2,966 tasks that do not reach full marks. The second round re-evaluates these candidates with GPT-5.5 (Codex, reasoning effort medium), removes tasks that are fully solved, and keeps 1,293. For non-binary scores such as F1, ``not fully solved'' means a score below full marks.

\textbf{Stage 2: skill pairing and effectiveness validation.} Candidate skills come from a locally curated skill set, GitHub repository search, the Bioconductor index, and skill marketplace listings, 15,368 entries after merging. Two rounds of filtering by domain relevance, documentation completeness, and maintenance status retain 1,996 entries, of which 1,154 are native agent skills; the retrieval pool actually used for pairing contains 1,478 entries. Pairing proceeds in three steps: the model first proposes 3--6 required capabilities from the task (without solving it), lexical retrieval over skill names and descriptions then keeps at most 25 candidates, and finally the model re-ranks them and returns 1--3 skills. The retriever is not given reference answers at this stage. Each candidate skill is then provided on its own, the agent is asked to read it and perform the relevant operations, and pairs that can be completed under Operationally Supported Skill Use are kept. Some BixBench skills were organized into procedural documents from public methodological material and re-validated for effectiveness.

\textbf{Stage 3: screening for an autonomous-use gap.} For each task--skill pair, we compare Natural Skill Use with Operationally Supported Skill Use using the same skill version, and retain tasks that are not completed under the former but completed under the latter. Screening primarily used GLM-5.1, with one comparison planned per task--skill pair. For the 113 retained tasks, the screening records contain 163 comparisons satisfying this criterion: 74 tasks have one such comparison, and 39 have multiple comparisons. The additional comparisons arose mainly from retries following API errors, program failures, or premature termination.

\textbf{Final composition.} The retained tasks derived from public benchmarks come from MedCalc-Bench and BixBench.

\begin{itemize}
\tightlist
\item
  \textbf{MedCalc-Bench (78 tasks).} These cover four calculators: HAS-BLED (8), PSI (2), Delta gap (30), and albumin-corrected Delta gap (38). The prompts are rewritten so that they do not name the calculator or the skill directly, while keeping the medical information needed for the calculation. Numerical variants are generated with the official task-expansion program. One of the 78 tasks is a neutrally worded version of an existing acid--base task.
\item
  \textbf{BixBench (35 tasks).} These are spread over 4 data capsules: multi-table integration and association analysis (bix-10, 18 tasks), paired matrix analysis (bix-16, 2), cohort and variant aggregation (bix-20, 6), and expression--feature association (bix-22, 9). Five are official original tasks; the other 30 construct new analysis slices, parameter conditions, or analysis conventions on the real data of the same capsule. Input data are mounted read-only, and the reference notebooks and answers are not visible to the agent.
\end{itemize}

\subsection{Authored and Expanded Tasks}

\textbf{Motivation and materials.} These tasks are inspired by the patterns of real user questions on a research-agent platform, but contain no original user records. The task materials come from public literature and data, on which the problem conditions are varied. The four task groups are constructed as follows (Table~\ref{tab:A2}).

\begin{table}[!htb]
\centering\small
\setlength{\belowcaptionskip}{6pt}
\caption{Materials and construction of the authored and expanded tasks. "Representative" is the number of tasks in EBU-Bench Lite.}\label{tab:A2}
\begin{tabularx}{\linewidth}{lrrX}
\toprule
Task group & Tasks & \makecell[r]{Representative} & Materials and construction \\
\midrule
Abstract screening & 61 & 21 & Real study abstracts and systematic-review settings from TrialPanorama; the inclusion criteria are changed and the target study is re-judged under the new criteria \\
Clinical evidence screening & 53 & 10 & As above \\
Variant functional evidence & 53 & 6 & Based on the ClinGen SVI PS3/BS3 functional-evidence framework, two tasks set against real functional-assay papers; given the validation panel, controls, and calibration conditions, the evidence code and strength are determined \\
Enrichment review & 51 & 4 & One task uses a real GEO gene signature (GSE10586, obtained via Enrichr) with GO Biological Process 2025 annotations; the assay panel and experimental design are specified, and the correct background set must be rebuilt before computing \\
\bottomrule
\end{tabularx}
\end{table}

\textbf{Literature screening tasks.} The 31 representative tasks of the two screening groups correspond to 20 studies and 18 systematic reviews, with some studies shared between the groups. The typical construction modifies the inclusion criteria of the review. For example, the original review allowed quasi-randomized studies and the target study used alternate allocation; once this allowance is removed, the study should be excluded. The reference answer is therefore the judgment under the modified criteria, not the label in the original dataset. Other versions change the abstract wording, joint eligibility conditions, or resource versions.

\textbf{Variant functional evidence and Enrichment review tasks.} The background knowledge of these two groups comes from public frameworks and data, while the validation records, assay panels, and experimental conditions in the tasks are controlled synthetic material that does not represent real patients or the experimental results of the original papers. Reference answers are computed by frozen rules: the variant tasks give the evidence code, strength, and number of usable validation variants; the enrichment tasks give the size and hash of the background set and, for each term, K, n, k, and the hypergeometric right-tail p value.

\textbf{Screening.} The authored tasks also pass through the autonomous-use gap screening.

\subsection{Scoring}

All tasks are scored only on the agent's final answer, without regard to whether the skill was opened or which tools were called. A passing answer scores 1 and any other answer 0; a missing answer or one whose format does not meet the scoring requirements scores 0.

\par\noindent\textbf{MedCalc-Bench.}
We use the official scoring function \texttt{check\_correctness}
with frozen numerical bounds.

\par\noindent\textbf{BixBench.}
We use the original scorer: 33 tasks are judged by numerical
intervals and 2 by string matching.

\par\noindent\textbf{Authored and expanded tasks.}
The 41 representative tasks share one scoring codebase,
with the rules in Table~\ref{tab:A3}.
\begin{table}[!htb]
\centering\small
\caption{Scoring rules of the authored and expanded tasks.}\label{tab:A3}
\begin{tabularx}{\linewidth}{>{\hsize=0.7\hsize}X>{\hsize=1.0\hsize}X>{\hsize=1.3\hsize}X}
\toprule
Task group & What the agent submits & Condition for a correct answer \\
\midrule
Abstract screening, Clinical evidence screening & An \texttt{included} or \texttt{excluded} judgment for the target study (identified by PMID) & The target PMID appears exactly once, the judgment matches the reference, and the submitted PMID set equals the expected set \\
Variant functional evidence & Evidence code (PS3/BS3/none), strength, number of usable validation variants & All three fields match the reference; integers must be exactly equal \\
Enrichment review & Size and hash of the background set; K, n, k, and p value for each term & All fields match; integers must be exactly equal, and floating-point values allow a relative error of 1e-5 \\
\bottomrule
\end{tabularx}
\end{table}

\subsection{Selection of Representative Tasks for EBU-Bench Lite}

EBU-Bench Lite contains 154 tasks: all 113 tasks derived from public benchmarks, plus one representative task from each ``source question--associated skill'' group of the 218 authored and expanded tasks, 41 in total (Table~\ref{tab:A4}). This keeps all 12 task groups while reducing the repeated weight of multiple variants of the same source question and skill in aggregate results.

\begin{table}[!htb]
\centering\small
\caption{Task composition of the full benchmark and EBU-Bench Lite.}\label{tab:A4}
\begin{tabularx}{\linewidth}{lrrX}
\toprule
Source or task group & \makecell[r]{Full\\benchmark} & \makecell[r]{EBU-Bench\\Lite} & Selection \\
\midrule
MedCalc-Bench-derived tasks & 78 & 78 & All retained \\
BixBench-derived tasks & 35 & 35 & All retained \\
Abstract screening & 61 & 21 & One per source question–skill combination \\
Clinical evidence screening & 53 & 10 & One per source question–skill combination \\
Variant functional evidence & 53 & 6 & One per source question–skill combination \\
Enrichment review & 51 & 4 & One per source question–skill combination \\
\textbf{Total} & \textbf{331} & \textbf{154} & — \\
\bottomrule
\end{tabularx}
\end{table}

\textbf{Grouping.} A ``source question'' is the original problem identifier used in the construction records to trace rewriting and expansion. Each task is keyed by its source question and the name of its associated skill: the same source question paired with different skills falls into different groups, whereas rewrites of the question or revisions of the resources under the same source question and skill fall into the same group. Grouping follows the existing construction relations, not text similarity or task scores. The 218 tasks form 41 groups of 1--40 tasks each; merging identical source questions across skills gives 31 source questions.

\textbf{Representative selection.} Each group selects one task by the following rules in order, stopping at the first rule that discriminates:

\begin{enumerate}
\def\labelenumi{\arabic{enumi}.}
\tightlist
\item
  By the source role in the construction records: tasks marked positive first, then tasks without a role label, and control tasks last.
\item
  Within the same priority, the task that comes first in the version order of the source index.
\item
  If still tied, the task whose directory name comes first in lexicographic order.
\end{enumerate}

\FloatBarrier
% Uses the existing manuscript preamble and refs_candidate.bib.
\subsection{Comparison with Related Skill Benchmarks}
\label{app:skill-benchmark-comparison}

Existing skill benchmarks evaluate the task gains from providing skills, the process of skill use, and reuse across instances. EBU-Bench builds on paired task-outcome evaluation and focuses on how much of the benefit of the same skill is realized: it holds the task and skill version fixed, compares Natural Skill Use with Operationally Supported Skill Use, and measures the task gains that remain unrealized during autonomous execution. Table~\ref{tab:related-skill-benchmarks} summarizes the domains, scale, and evaluation designs of five related benchmarks and EBU-Bench. The comparison below explains how successful-use validation, outcome differences, and paired-trace analysis support the diagnosis of skill underuse.

\begin{table}[!htb]
\centering
\caption{Comparison of evaluation designs in related skill benchmarks.}
\label{tab:related-skill-benchmarks}
\begingroup
\footnotesize
\setlength{\tabcolsep}{3pt}
\renewcommand{\arraystretch}{1.16}
\begin{tabularx}{\linewidth}{@{}>{\raggedright\arraybackslash}p{0.17\linewidth}
>{\hsize=0.90\hsize\linewidth=\hsize}X
>{\hsize=1.25\hsize\linewidth=\hsize}X
>{\hsize=1.10\hsize\linewidth=\hsize}X
>{\hsize=0.90\hsize\linewidth=\hsize}X
>{\hsize=0.85\hsize\linewidth=\hsize}X@{}}
\toprule
Benchmark & Domain / scale & Conditions & Representative metric & Skill-access analysis & Successful-use validation \\
\midrule
SkillsBench & 8 domains; 87 tasks & No skills / mounted skills & Mean verifier score and skill gain & Yes: invocation rate & --- \\
\addlinespace[3pt]
SWE-Skills-\allowbreak Bench & 6 SWE subdomains; 980 requirements & No skill / forced loading & Gain in executable-check pass fraction & --- & --- \\
\addlinespace[3pt]
Skill-Use & 9 domains; 177 tasks; 79 skills & On-demand reading / full preloading & Skill-use process score (SU) & Yes: triggering and compliance & --- \\
\addlinespace[3pt]
OpenSkillEval & 5 artifact types; 677 instances; 30 skills & No skills / forced use & Artifact-quality scores & Yes: reading rate & --- \\
\addlinespace[3pt]
SkillLearnBench & 15 subdomains; 20 tasks; 100 instances & No skills / human skills / generated skills & Task accuracy across instances & Yes: usage rate and trace alignment & Yes: human-authored skill sets \\
\addlinespace[3pt]
\textbf{EBU-Bench (ours)} & \textbf{Medical and biomedical; 331 tasks (Lite: 154)} & \textbf{No skill / natural / prompted / supported use} & \textbf{Supported--natural accuracy gap} & \textbf{Yes: opening rate and trace analysis} & \textbf{Yes} \\
\bottomrule
\end{tabularx}
\par\smallskip
\raggedright
\textit{Notes.} ``Yes'' indicates that the cited version reports the corresponding evaluation or analysis; ``---'' indicates that it is not reported. Skill-access analysis includes access records and related process measures. Successful-use validation means executing the task with the paired skill during construction and requiring successful completion for inclusion. Scale follows each paper's own unit: task, requirement, or instance.
\par\smallskip
SkillsBench has been accepted at NeurIPS 2026; this comparison uses arXiv:2602.12670v4 (87 tasks, versus 84 in the conference-page abstract). SWE-Skills-Bench and Skill-Use are preprints; we use arXiv:2603.15401v2 and arXiv:2608.04828, respectively. For OpenSkillEval we use the Findings of EMNLP 2026 version, and for SkillLearnBench the COLM 2026 version.\par
\endgroup
\end{table}

\textbf{Successful-use validation establishes an empirical reference for diagnosis.}
Related benchmarks use different construction checks to establish the relationship between tasks and skills. SkillsBench combines end-to-end oracle verification with actual with/without-skill runs: the reference solution must pass all tests, tasks must be tested with agents under both conditions, and tasks are screened for measurable skill-related gains (\citealp{li2026skillsbenchbenchmarkingagentskills}, \S3 and Appendices B.4--B.7). Whether the oracle must use the paired skills, and whether every task must be fully solved in the with-skill condition, are not specified. SWE-Skills-Bench manually screens a subset of public skills and explicitly loads the designated skill in its paired evaluations \citep{han2026sweskillsbenchagentskillsactually}. Skill-Use validates the executability, cross-component consistency, and scoring reliability of skill--task--rubric triples \citep{han2026skillusellmsactuallyuse}. During construction, OpenSkillEval checks that task specifications and instructions are information-complete and consistent with the source materials, filtering inconsistent instances; subsequent skill-augmented executions compare final artifact quality (\citealp{ying2026openskillevalautomaticallyauditingopen}, \S\S2.1 and 3.1). SkillLearnBench uses the same or a stronger model to confirm at least one successful completion per instance with the human-authored skill set (\citealp{zhong2026skilllearnbench}, \S3.1); this construction check applies to human-authored reference skills, while generated skills are assessed in the subsequent evaluation.

\textbf{EBU-Bench requires successful task completion with the paired skill under operational support for inclusion, establishing execution evidence of successful use for each retained task--skill pair.} It then retains tasks that are not completed under Natural Skill Use, focusing on cases with demonstrated successful use but unrealized gains during autonomous execution. Beyond checking task specifications and source-material consistency, this validation directly tests whether the paired skill can support completion of the current task, providing a successful-use reference for the subsequent natural-use evaluation. The resulting challenge set supports targeted study of how skill benefits are realized across models and harnesses.

\textbf{Measuring the gap from providing a skill to realizing its benefits.}
SkillsBench lets agents discover and use one or more mounted skills without naming them in the task, and measures their gains through with/without-skill comparisons. SkillLearnBench further compares human-authored and generated skills across instances of the same task. SWE-Skills-Bench loads the full skill through a native command, while OpenSkillEval explicitly requires skill use; both compare against a no-skill condition. Skill-Use holds skill content fixed in its on-demand-reading versus full-preloading experiment, pairing runs by task and model under Codex to study the effect of access conditions, and additionally tests distractor skills.

EBU-Bench combines comparisons of use conditions for the same skill with final task outcomes. It defines four conditions for each task; the three skill-provided conditions---Natural, Prompted, and Operationally Supported Skill Use---share the same skill version, with one matched skill per task. Under Natural Skill Use, the agent decides whether to read the skill and how to execute it. Operationally Supported Skill Use explicitly requires reading the skill and performing its task-relevant operations. No Skill and Prompted Skill Use provide references for performance without the skill and the effect of a general reminder. Skill-use accuracy is the mean binary score of final answers, and the EBU gap is
\(\Delta_{\mathrm{EBU}}(m,h)=P_{\mathrm{supported}}(m,h)-P_{\mathrm{natural}}(m,h)\),
where \(m\) and \(h\) denote the model and harness, and \(P\) is the mean task score under the corresponding condition; the main evaluation uses 154 Lite tasks. Performance under operational support thus provides an empirical reference, and the EBU gap quantifies task gains that remain unrealized under natural use.

\textbf{The EBU-Bench metric extends task-performance evaluation to the diagnosis of unrealized skill benefits.} SkillsBench, SWE-Skills-Bench, OpenSkillEval, and SkillLearnBench assess skill effects through verifier scores, executable checks, artifact quality, or task accuracy; Skill-Use's SU score focuses on triggering, compliance, and boundary constraints. EBU-Bench instead uses supported-use accuracy on the same tasks and skills as an empirical reference for the gains left unrealized under autonomous use. \textbf{Even when providing a skill already improves performance, the evaluation can reveal how much additional benefit remains unrealized.} Paired execution traces connect this gap to skill access, rule acquisition, and rule application, informing revisions to skill descriptions and execution guidance.

\textbf{Paired-trace analysis connects skill access with task gains.}
Existing benchmarks provide several process-level observations. SkillsBench reports invocation rates and performance on the invoked subset (Appendix K, Table 14). Skill-Use gates SU by triggering and compares triggering rates, compliance, and boundary adherence under on-demand reading and full preloading. Before its main experiments, OpenSkillEval reports reading rates of 48\% and 94\% under default and forced-use conditions. SkillLearnBench reports trajectory alignment and skill usage rate alongside accuracy, while SWE-Skills-Bench reports behavioral counts in selected skill-level case studies in \S4.2.

\FloatBarrier

\clearpage
% generated by build_en_fable.py from ../en_fable_md/9_appendix_B.md -- do not edit by hand
\section{Method Details of EBU-Evolve}

This appendix gives the reflection inputs, candidate construction, and selection rules of EBU-Evolve as shown in Figure~\ref{fig:overview}c.

\subsection{Paired Evidence and Reflection Inputs}

For the original skill \(s_0\), we run Natural Skill Use and Operationally Supported Skill Use under the same model, harness, and execution limits. Both sides receive the same task inputs and skill resources; the latter additionally appends a predetermined operational-support prompt. The prompt requires reading or applying the existing resources and does not supply the answer to the current task. For the tasks authored around research questions, the prompt requires reading the skill entry file and consulting the necessary reference materials it points to.

The reflection model receives the following materials:

\begin{table}[!htb]
\centering\small
\begin{tabularx}{\linewidth}{lX}
\toprule
Material & Content and use \\
\midrule
Original skill and reference materials & Provide the rules, applicability conditions, and operational basis for the revision \\
Overview of all training pairs & Records both sides' task scores and access facts, showing improved, unchanged, and degraded pairs \\
Selected complete pairs & Keep the prompt, tool calls and returns, final answers, task scores, and access records for analyzing specific changes \\
Training access diagnosis & Distinguishes instructions not read, instructions read but misapplied, and access status unknown \\
\bottomrule
\end{tabularx}
\end{table}

When selecting complete pairs, we prioritize cases whose Natural Skill Use score improves, include unchanged or degraded cases when the input limit allows, and fill the remainder in a predetermined order. If the context budget is exceeded, the number of cases is reduced while every selected trace is kept in full; the overview of all training pairs is still supplied alongside the complete cases. The overview may be cited for scores and access facts, whereas specific revisions must be supported by complete cases and the original materials. The reflection model maps the differences observed in the paired executions to existing rules; pairs without improvement are used to test the scope of the proposed revisions.

Access status is judged from explicit operations in the execution records, not inferred from answer correctness or the appearance of file names; it is left unknown when records are missing. If the agent did not read the relevant instructions, the reflection model considers whether the description needs adjusting; if it did, the model further examines how the rules were applied.

The reflection output includes the pairs supporting the change, counterexamples to consider, citations to the original materials, reusable revision rules, and suggested locations. Revision rules state which applicability conditions or execution steps should be clarified, moved earlier, or reorganized, covering a rule's applicability conditions, execution steps, and how its results are used. Citations must be locatable in the original skill package. Neither reflection nor editing receives per-task prompts, answers, traces, or scores from the validation or test sets.

\subsection{Candidate Construction with a Shared Body}

Revision rules take effect only if the agent reads and correctly applies them during autonomous execution; candidates therefore consider both the description and the body.

Let the original skill be \(s_0=(d_0,b_0)\), with description \(d_0\) and body \(b_0\). From the original materials, distilled revision rules, and training utilization diagnosis, the editing model produces a revised description \(d'\) and body \(b'\). We construct three candidates:
\[
s_{\mathrm{body}}=(d_0,b'),\qquad
s_{\mathrm{joint}}=(d',b'),\qquad
s_{\mathrm{description}}=(d',b_0).
\]
The body-only and joint candidates share the revised body; the joint and description-only candidates share the revised description. This construction compares editing scopes from the same edit. All candidates preserve the other metadata and reference resources and are reviewed separately. Those passing review enter validation-set selection.

The description tells the executing agent when the skill applies and which requirements are essential, while the body organizes the complete rules and steps. Added content must preserve the applicability conditions of the original method and must not store training-task identifiers, specific values, or answers.

\subsection{Candidate Review}

Editing is limited to the body of \texttt{SKILL.md} and the description where permitted; other metadata, reference files, code, and data remain unchanged. Every candidate is subject to an edit-size cap, computed as the number of lines added plus lines deleted relative to the parent version, where replacing a line counts as one deletion and one addition. Passages designated for preservation must remain locatable in the candidate.

A source check verifies newly added identifiers, existing dependencies, and links to reference files. An independent semantic review further checks whether the changes are supported by the original materials, whether the applicability conditions of a rule have been altered incorrectly, and whether the description still addresses the executing agent. Proposals that fail the review are not evaluated.

A candidate that fails the review may be repaired once according to the review comments; the repair does not use feedback from the validation or test sets.

\subsection{Validation-Set Selection}

Selection is performed under Natural Skill Use, reproducing how a skill is read and executed by the agent on its own in practice.

The original skill and the eligible candidates are evaluated under Natural Skill Use on the same validation set, with the same model, harness, and per-task execution limits. A candidate must complete all scheduled tasks and repetitions with a valid score each time; candidates with incomplete evaluations do not enter the selectable set, and missing scores are not filled with zeros.

To stabilize selection, each task is repeated \(K\) times (\(K=3\) unless otherwise specified); scores are first averaged within a task and then across tasks:

\[
J(s)=\frac{1}{|\mathcal V|}\sum_{i\in\mathcal V}
\left[\frac{1}{K}\sum_{k=1}^{K}y_{i,k}(s)\right].
\]

Versions are ranked by \(J(s)\). A tie with the original keeps the original; a tie between revisions prefers the one with the smaller edit relative to the original, then a fixed candidate order. A revision is adopted only if its score strictly exceeds the original. The original always takes part in the comparison, so that no revision is adopted without an actual improvement. Skill-opening rate, process predicates, and the EBU gap do not enter the selection score.

\subsection{Budget and Algorithm}

Each search fixes in advance the call limits for execution, reflection, editing, and review, and records the actual usage of each. One edit yields three candidates, and validation evaluations of each count against the execution budget. The final test is accounted separately, and test feedback is never used to generate or select candidates. EBU-Evolve performs one round of paired diagnosis and candidate selection on the original skill. The algorithm is as follows:

\begin{algorithm}[H]
\caption{EBU-Evolve: paired diagnosis and skill revision}
\label{alg:ebu-evolve}
\small
\begin{algorithmic}[1]
\Require Original skill $s_0=(d_0,b_0)$ and its reference materials
\Require Training set $\mathcal T$, validation set $\mathcal V$,
         and test set $\mathcal U$
\Require Fixed operational-support prompt and search budget
\Ensure Frozen skill $s^\star$ and held-out test results

\State Evaluate $s_0$ on $\mathcal V$ under Natural Skill Use
       to obtain $J(s_0)$.
\State Initialize the selectable set $\mathcal C \gets \{s_0\}$.
\State Collect paired executions of $s_0$ on $\mathcal T$ under
       Natural Skill Use and Operationally Supported Skill Use.
\State Reflect on complete pairs, the training overview, and original
       materials to derive reusable revision rules.

\If{the evidence supports a reusable revision}
    \State Generate a revised body $b'$ and description $d'$.
    \State Construct $s_{\mathrm{body}}=(d_0,b')$, $s_{\mathrm{joint}}=(d',b')$,
           and $s_{\mathrm{description}}=(d',b_0)$.
    \ForAll{$s \in \{s_{\mathrm{body}},s_{\mathrm{joint}},s_{\mathrm{description}}\}$}
        \State Apply the structural, source, and semantic checks
               and the repair rule in Appendix~B.3.
        \If{$s$ passes review}
            \State Evaluate $s$ on $\mathcal V$ under Natural Skill Use
                   with $K$ repetitions per task (Appendix~B.4).
            \If{all scheduled evaluations have valid scores}
                \State Compute $J(s)$ as defined in Appendix~B.4.
                \State $\mathcal C \gets \mathcal C \cup \{s\}$.
            \EndIf
        \EndIf
    \EndFor
\EndIf

\State Select $s^\star$ from $\mathcal C$ by the highest $J(s)$,
       using the tie-breaking rules in Appendix~B.4.
\State Keep $s_0$ unless a candidate strictly improves on $J(s_0)$.
\State Freeze $s^\star$ and compare it with $s_0$ on $\mathcal U$
       under Natural Skill Use.
\State Do not use test feedback for revision or selection.
\end{algorithmic}
\end{algorithm}
\subsection{Final Evaluation and Information Use}

Training, validation, and test sets are split in advance by task ID and used, respectively, to generate revisions, select the version, and evaluate the final skill. The selected version is handed to the agent as an ordinary skill; later tasks carry no training traces, reflection records, or operational-support prompts. Cross-model and cross-harness evaluations use the same text and compare against the original skill in the target setting; target test results are not used to re-edit or re-select the version.

\subsection{Data Splits and Execution Settings}

The numbers of training, validation, and test tasks of each skill are given in Table~\ref{tab:B1}; the three source models use the same split, except that the Enrichment review training and validation sets of GPT-5.5 contain 14 and 10 tasks. Within a target configuration, the original and evolved skills use the same execution settings; the output limit of a single run is 49,152 tokens (64,000 for Atria). Runs interrupted by infrastructure failures are rerun according to the protocol.

\begin{table}[!htb]
    \centering
    \caption{Task splits of each skill (number of tasks).}
    \label{tab:B1}

    \small
    \begin{tabular*}{\linewidth}{@{\extracolsep{\fill}}lrrr@{}}
        \toprule
        Skill & Training & Validation & Test \\
        \midrule
        Abstract screening          & 16 & 16 & 20 \\
        Clinical evidence screening & 22 & 18 & 13 \\
        Enrichment review           & 11 & 13 & 27 \\
        Variant functional evidence & 13 & 13 & 27 \\
        \midrule
        \textbf{Total}
        & \textbf{62}
        & \textbf{60}
        & \textbf{87} \\
        \bottomrule
    \end{tabular*}
\end{table}

\subsection{Ablations}

\begin{figure}[!htb]\centering
\begin{minipage}[b]{0.49\textwidth}\centering
\includegraphics[width=\linewidth,trim={0pt 3pt 877.12pt 44pt},clip]{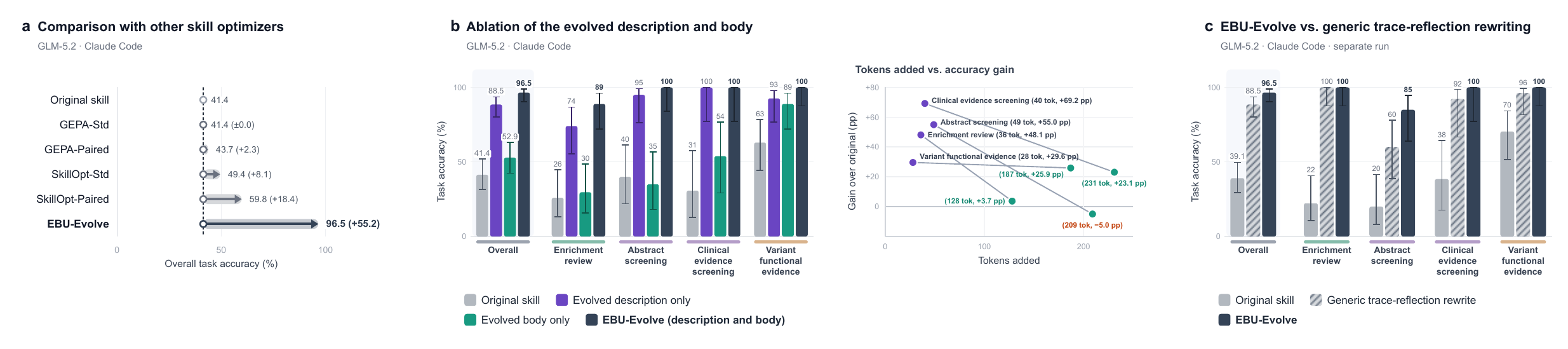}\\[-1pt]
{\small (a) Comparison with GEPA and SkillOpt}
\end{minipage}\hfill
\begin{minipage}[b]{0.49\textwidth}\centering
\includegraphics[width=\linewidth,trim={883pt 3pt 0pt 44pt},clip]{figures/ebu-evolve-ablation-v7_noframe.pdf}\\[-1pt]
{\small (c) Comparison with generic trace-reflection rewriting}
\end{minipage}\\[8pt]
\includegraphics[width=\textwidth,trim={308pt 3pt 302.12pt 44pt},clip]{figures/ebu-evolve-ablation-v7_noframe.pdf}\\[-1pt]
{\small (b) Decomposition into description and body}
\caption{Method comparison and ablations, all evaluated under GLM-5.2/Claude Code. (a) Comparison of the frozen artifacts with the prompt-optimization and skill-optimization methods; (b) ablation of the presentation components of the frozen GLM-5.2 artifact, where the scatter plot on the right shows the added text and test gain of the description-only and body-only revisions, with tokens counted uniformly by o200k\_base; (c) comparison of EBU-Evolve with generic trace-reflection rewriting, run as a separate batch and not pooled with (a) or (b). Error bars are per-bar Wilson 95\% intervals.}\label{fig:ebu-evolve-ablation-v7}
\end{figure}

\subsubsection{Decomposition into Description and Body}

Let the frozen joint version of GLM-5.2 be \(s_{DB}=(d',b')\); we split off the description-only version \(s_D=(d',b_0)\) and the body-only version \(s_B=(d_0,b')\) and compare them with the original \(s_0=(d_0,b_0)\). \(s_D\) and \(s_B\) retain, respectively, the description and the body actually revised in the joint version.

On the frozen GLM-5.2 artifact, the joint revision attains the highest skill-use accuracy, 96.55\%; the description-only revision reaches 88.51\% and the body-only revision 52.87\% (Table~\ref{tab:B2}, Figure~\ref{fig:ebu-evolve-ablation-v7}b). The description-only revision retains 85.42\% of the net gain of the joint revision over the original. It outperforms the body-only revision on all four skills while adding only 28--49 tokens, compared with 128--231 tokens for the body revision (Table~\ref{tab:B3}), indicating that a small revision of the entry point can already bring a large Natural Skill Use gain. The complementary effect of the body is more visible when combined with the new description: in Enrichment review, revising the body alone raises skill-use accuracy from 25.93\% to 29.63\%, a gain of 3.70 pp, whereas adding the same body on top of the new description raises it from 74.07\% to 88.89\%, a gain of 14.81 pp.~This suggests that the effect of a body revision should be assessed together with the description: the description may guide the agent to read further or may itself supply the key rule.

\begin{table}[!htb]
    \centering
    \caption{Accuracy (\%) of the decomposed versions under Natural Skill Use.}
    \label{tab:B2}

    \small
    \begin{tabular*}{\linewidth}{@{\extracolsep{\fill}}lrrrr@{}}
        \toprule
        Skill & Original & Description only & Body only & Joint \\
        \midrule
        Abstract screening
        & 40.00 & 95.00 & 35.00 & 100.00 \\
        Clinical evidence screening
        & 30.77 & 100.00 & 53.85 & 100.00 \\
        Enrichment review
        & 25.93 & 74.07 & 29.63 & 88.89 \\
        Variant functional evidence
        & 62.96 & 92.59 & 88.89 & 100.00 \\
        \midrule
        \textbf{Overall}
        & \textbf{41.38}
        & \textbf{88.51}
        & \textbf{52.87}
        & \textbf{96.55} \\
        \bottomrule
    \end{tabular*}
\end{table}

\begin{table}[!htb]
    \centering
    \caption{Length of the full \texttt{SKILL.md} (o200k\_base tokens).}
    \label{tab:B3}

    \small
    \begin{tabular*}{\linewidth}{@{\extracolsep{\fill}}lrrrr@{}}
        \toprule
        Skill & Original & Description only & Body only & Joint \\
        \midrule
        Abstract screening          & 89 & 138 & 298 & 347 \\
        Clinical evidence screening & 84 & 124 & 315 & 355 \\
        Enrichment review           & 91 & 127 & 219 & 255 \\
        Variant functional evidence & 66 &  94 & 253 & 281 \\
        \bottomrule
    \end{tabular*}
\end{table}
The description-only editing condition above retains access to the body and references. We use three mechanism comparisons to distinguish support from the remaining resources, the content foregrounded in the description, and actual skill access. Each comparison covers the same four task groups and 87 test tasks under Natural Skill Use, with its own frozen artifacts and evaluations. Results are compared within each experiment.

\paragraph{Resources beyond an unchanged description.}
We freeze one set of evolved skill artifacts and retain their descriptions, comparing access to the complete resources with a condition in which the body and reference files are inaccessible. Each condition is evaluated independently three times on the same tasks. The description provides the always-visible capability, applicability, and entry information; the body and references provide the fuller procedures and decision criteria.
\begin{table}[!htbp]
    \centering
    \caption{Resource availability with the description fixed. Conditions are evaluated three times. Accuracies are reported as three-evaluation means, with counts expressed on the task-group denominator.}
    \label{tab:B4}

    \small
    \begin{tabular*}{\linewidth}{@{\extracolsep{\fill}}lrr@{}}
        \toprule
        Task group & Complete resources & Description visible only \\
        \midrule
        Abstract screening
            & 85.00\% (17/20) & 55.00\% (11/20) \\
        Clinical evidence screening
            & 84.62\% (11/13) & 61.54\% (8/13) \\
        Enrichment review
            & 100.00\% (27/27) & 59.26\% (16/27) \\
        Variant functional evidence
            & 92.59\% (25/27) & 33.33\% (9/27) \\
        \midrule
        Overall
            & 91.95\% (80/87) & 50.57\% (44/87) \\
        \bottomrule
    \end{tabular*}
\end{table}

\begin{figure}[!tb]
    \centering
    \includegraphics[
        width=0.88\linewidth
    ]{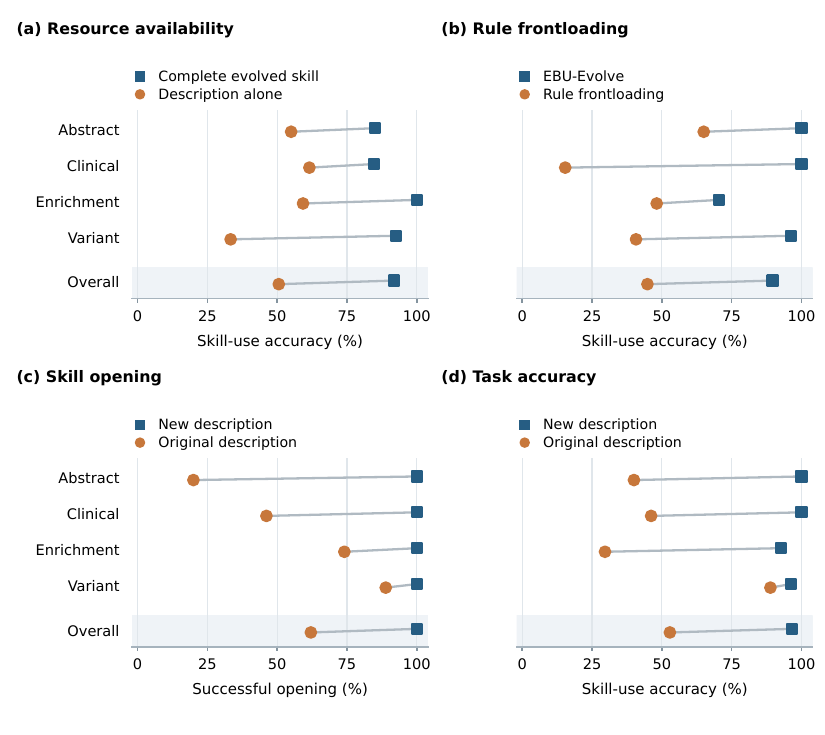}
    \caption{Mechanism comparisons within Appendix~B.8.1. (a) Resource availability with the description fixed. (b) Rule frontloading from original materials versus EBU-Evolve. (c--d) Opening and accuracy in the independent description-access experiment, with the body and references fixed. Panels (a), (c), and (d) report means over three evaluations; the conditions and frozen artifacts are defined within their respective experiments.}
    \label{fig:description-mechanisms-b81}
\end{figure}

Accuracy is 91.95\% (80/87) with complete resources and 50.57\% (44/87) with only the descriptions visible, a difference of 41.38 pp. All four groups decrease, by 6, 3, 11, and 16 correct answers for Abstract screening, Clinical evidence screening, Enrichment review, and Variant functional evidence, respectively. These results show that the descriptions do not replace the complete resources on this test panel (Table~\ref{tab:B4}, Figure~\ref{fig:description-mechanisms-b81}a).

\paragraph{Rule frontloading from the original materials.}
The rewriting model receives only the original skill and its references, without execution traces or utilization diagnoses. The proposed text is incorporated into the original description after length constraints, source review, and validation-set selection. We evaluate this rule-frontloading condition alongside EBU-Evolve artifacts; validation scores are used for selection rather than supplied as execution traces to the generator.
\begin{table}[!htbp]
    \centering\small
    \caption{Rule frontloading without execution feedback during generation, compared with EBU-Evolve. Entries give task accuracy and correct-answer counts.}
    \label{tab:B5}

    \begin{tabular*}{\linewidth}{@{\extracolsep{\fill}}lrr@{}}
        \toprule
        Task group & Rule frontloading & EBU-Evolve \\
        \midrule
        Abstract screening
            & 65.00\% (13/20) & 100.00\% (20/20) \\
        Clinical evidence screening
            & 15.38\% (2/13) & 100.00\% (13/13) \\
        Enrichment review
            & 48.15\% (13/27) & 70.37\% (19/27) \\
        Variant functional evidence
            & 40.74\% (11/27) & 96.30\% (26/27) \\
        \midrule
        Overall
            & 44.83\% (39/87) & 89.66\% (78/87) \\
        \bottomrule
    \end{tabular*}
\end{table}

Rule frontloading achieves 44.83\% (39/87), compared with 89.66\% (78/87) for EBU-Evolve (Table~\ref{tab:B5}, Figure~\ref{fig:description-mechanisms-b81}b). The difference is largest in Clinical evidence screening, at 15.38\% versus 100.00\%. In this task group, the frontloaded description emphasizes general screening requirements, whereas the evolved description directs the agent to judge study design from the actual allocation method. The comparison highlights the importance of which rules and applicability conditions are foregrounded, beyond placing rules earlier in the description.

\paragraph{Description revision and skill opening: an independent experiment.}
We conduct a separate description-access experiment using a frozen set of evolved skills. Within this experiment, the body, reference files, and remaining metadata are held fixed while the description is switched between its original and evolved forms. The two conditions use independent sessions in randomized alternating order, with three executions per task and condition. We measure opening and correctness from the same runs. This experiment has its own evaluation records, distinct from the revision-location comparison above.
Opening follows the successful-reading criterion in Section~5.1 and Appendix~E.1. A successful read of the target skill or a successful native load counts as opening; directory listings, path mentions, and failed reads do not. All recorded openings in this experiment have successful-reading evidence.
\begin{table}[!htbp]
    \centering\small
    \caption{Independent description-access experiment with the body and reference files fixed. Each task and condition is evaluated in three independent sessions. Counts are expressed on the task-group denominator for the reported mean rates. Opening and correctness are scored separately.}
    \label{tab:B6}

    \begin{tabular*}{\linewidth}{@{\extracolsep{\fill}}lrrrr@{}}
        \toprule
        Task group
            & \makecell{Original\\description\\opened}
            & \makecell{Evolved\\description\\opened}
            & \makecell{Original\\description\\correct}
            & \makecell{Evolved\\description\\correct} \\
        \midrule
        Abstract screening
            & 4/20 & 20/20 & 8/20 & 20/20 \\
        Clinical evidence screening
            & 6/13 & 13/13 & 6/13 & 13/13 \\
        Enrichment review
            & 20/27 & 27/27 & 8/27 & 25/27 \\
        Variant functional evidence
            & 24/27 & 27/27 & 24/27 & 26/27 \\
        \midrule
        Overall
            & 54/87 & 87/87 & 46/87 & 84/87 \\
        \bottomrule
    \end{tabular*}
\end{table}
The opening rate increases from 62.07\% (54/87) to 100\% (87/87), a gain of 37.93 pp. Task accuracy increases from 52.87\% (46/87) to 96.55\% (84/87), a gain of 43.68 pp. Both measures increase in all four task groups (Table~\ref{tab:B6}, Figure~\ref{fig:description-mechanisms-b81}c--d). With the body and references fixed, the description intervention provides direct support for improved skill opening and shows improved task performance under Natural Skill Use. Opening and task correctness remain separate outcomes: this experiment measures the effect of description revision on each, rather than the proportion of the accuracy gain mediated by reading.

\subsubsection{Comparison with Generic Trace-Reflection Rewriting}

The setup of this control is described in §4.4; it is run as a separate batch under GLM-5.2/Claude Code, so its original-skill accuracies differ from those in Figures~\ref{fig:ebu-evolve-main-v4} and \ref{fig:ebu-evolve-ablation-v7}a. Both groups receive the original skill and the complete reference materials; the comparison concerns how revision evidence is obtained from execution. EBU-Evolve uses the paired executions of the same task under Operationally Supported Skill Use to locate where an existing rule was not acquired or not carried out, and writes the corresponding requirement back into the skill. Overall, EBU-Evolve answers 7 more of the 87 tasks correctly, a skill-use accuracy 8.05 pp higher (96.55\% vs.~88.51\%); 5 of these come from Abstract screening (85.00\% vs.~60.00\%), contributing 5.75 pp, while Clinical evidence screening (100.00\% vs.~92.31\%) and Variant functional evidence (100.00\% vs.~96.30\%) each contribute one more correct answer, and both groups reach 100.00\% on Enrichment review (Table~\ref{tab:B7}, Figure~\ref{fig:ebu-evolve-ablation-v7}c). The results support the value of EBU-Evolve's overall EBU-driven design.

\begin{table}[!htb]
    \centering
    \caption{Per-skill comparison with generic trace-reflection rewriting
    (accuracy under Natural Skill Use, \%).}
    \label{tab:B7}

    \small
    \begin{tabular*}{\linewidth}{@{\extracolsep{\fill}}lrrr@{}}
        \toprule
        Skill
        & Original
        & \makecell[r]{Generic trace-reflection\\rewriting}
        & EBU-Evolve \\
        \midrule
        Abstract screening
        & 20.00 & 60.00 & 85.00 \\
        Clinical evidence screening
        & 38.46 & 92.31 & 100.00 \\
        Enrichment review
        & 22.22 & 100.00 & 100.00 \\
        Variant functional evidence
        & 70.37 & 96.30 & 100.00 \\
        \midrule
        \textbf{Overall}
        & \textbf{39.08}
        & \textbf{88.51}
        & \textbf{96.55} \\
        \bottomrule
    \end{tabular*}
\end{table}
\subsubsection{Paired Evidence, Case Selection, and Utilization Diagnosis}
\label{app:components-stability}
\label{app:component-contributions}

Appendix~B.8.2 compares full EBU-Evolve with generic trace-reflection rewriting. We now examine how paired execution evidence, gain-based case selection, and utilization diagnosis contribute to revision quality, and whether the advantage persists across optimization runs and task groups. We use GLM-5.2 under Claude Code and evaluate Natural Skill Use accuracy on the same 87 test tasks across four task groups.

We distinguish three components of the reflection input. \textbf{P (Paired Evidence)} provides traces of the same training task under Natural Skill Use and Operationally Supported Skill Use, allowing the rewriting model to compare their execution differences. \textbf{S (Gain-based Case Selection)} prioritizes training pairs whose scores improve under operational support, concentrating the limited reflection input on cases that may expose skill-utilization problems. \textbf{D (Utilization Diagnosis)} uses execution records to distinguish problems such as missing skill access and incorrect application after access, giving the revision a concrete target. S selects training cases for reflection; the final skill version is still selected by validation scores as described in Appendix~B.4.

All conditions fix the original skills, references, data splits, candidate construction and review procedures, and validation-set selection rules, while controlling trajectory-input and editing budgets. Generic Rewrite performs generic trace-reflection rewriting with basic revision instructions. Because gain-based selection depends on paired scores, we progressively add P, S, and D, and include P+D to compare selection and diagnosis on top of the same paired-evidence component. Each of the five conditions is independently optimized three times.

\begin{table}[!htbp]
    \centering\small
    \caption{Component ablations on 87 test tasks, with three independent optimization repeats. Filled circles indicate enabled components. Mean accuracy is computed from the unrounded task counts.}
    \label{tab:B8}

    \begin{tabular*}{\linewidth}{@{\extracolsep{\fill}}lcccrrrr@{}}
        \toprule
        Method & P & S & D & R1 & R2 & R3 & \makecell{Mean\\accuracy} \\
        \midrule
        Generic Rewrite
            & -- & -- & -- & 74/87 & 71/87 & 73/87 & 83.52\% \\
        + Paired
            & $\bullet$ & -- & -- & 75/87 & 74/87 & 76/87 & 86.21\% \\
        + Paired + Selection
            & $\bullet$ & $\bullet$ & -- & 78/87 & 76/87 & 78/87 & 88.89\% \\
        + Paired + Diagnosis
            & $\bullet$ & -- & $\bullet$ & 79/87 & 78/87 & 80/87 & 90.80\% \\
        Full EBU-Evolve
            & $\bullet$ & $\bullet$ & $\bullet$
            & \textbf{82/87} & \textbf{81/87} & \textbf{83/87}
            & \textbf{94.25\%} \\
        \bottomrule
    \end{tabular*}
\end{table}

\paragraph{Paired evidence provides a same-task reference for revision.}
Adding paired execution evidence to generic trace-reflection rewriting increases mean accuracy from 83.52\% to 86.21\%, a gain of 2.68 pp (Table~\ref{tab:B8}). Compared with reflecting on natural execution alone, traces of the same task under natural execution and operational support help the rewriting model identify which access or application steps change and whether those changes accompany a higher task score. This result supports paired traces as a more targeted basis for revision.

\paragraph{Case selection and utilization diagnosis further improve revision quality.}
Given paired evidence, prioritizing cases whose scores improve under operational support raises mean accuracy to 88.89\%, a further gain of 2.68 pp. Adding diagnosis of skill-access and application problems instead raises accuracy to 90.80\%, a gain of 4.60 pp. Gain-based selection focuses reflection on more informative execution differences, while utilization diagnosis helps determine whether the revision should address the reading entry point or the way a rule is applied. Diagnosis gives the larger improvement in this comparison, suggesting that identifying where the utilization problem occurs remains useful after execution contrasts are provided.

\begin{figure}[!ht]
    \centering
    \includegraphics[
        width=\textwidth
    ]{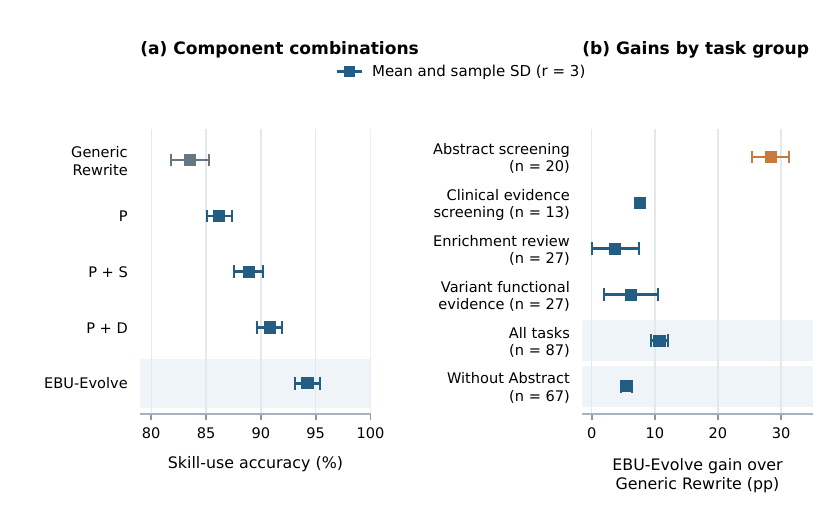}
    \caption{Component contributions and the distribution of gains across three independent optimization repeats. (a) Skill-use accuracy for the five component configurations. (b) EBU-Evolve's accuracy gain over Generic Rewrite within each task group, across all 87 tasks, and after excluding Abstract screening. Squares and error bars show the mean and sample standard deviation across three independent optimization repeats. The shaded rows highlight EBU-Evolve in (a) and the two task aggregates in (b).}
    \label{fig:components-stability}
\end{figure}

Full EBU-Evolve combines paired evidence, gain-based selection, and utilization diagnosis, reaching 94.25\% and achieving the highest correct-answer count in all three runs (Figure~\ref{fig:components-stability}a). Adding diagnosis to paired evidence with case selection improves accuracy by 5.36 pp; adding case selection to paired evidence with diagnosis improves it by 3.45 pp. These comparisons show that both choosing which cases to reflect on and interpreting their utilization problems contribute to revision quality. Their combination makes fuller use of same-task execution differences.

\subsubsection{Performance across Independent Optimization Runs}
\label{app:optimization-stability}

To examine whether the advantage depends on a particular generated and selected skill, we independently repeat the complete optimization procedure three times ($R=3$), starting from the same original skills and keeping the training, validation, and test splits fixed. Each run repeats trace reflection, editing, candidate review, and validation-set selection. Validation follows Appendix~B.4, with three executions per task ($K=3$) and selection by mean task score. Here, $R$ counts independent optimization runs, and $K$ counts validation executions within each run.

Each selected version is frozen before evaluation on the 87 test tasks. Test feedback does not enter candidate generation or selection. Table~\ref{tab:B9} expands the Generic Rewrite and full EBU-Evolve rows of Table~\ref{tab:B8}, using the same three runs to examine both mean gains and variation. Percentage-point differences use unrounded counts, and standard deviations are sample standard deviations across the three runs.

\begin{table}[!htbp]
    \centering\small
    \caption{Test performance across three independent optimization repeats. Accuracy and its standard deviation are expressed in percentages and percentage points, respectively; the standard deviation of net correct answers is in task counts.}
    \label{tab:B9}

    \begin{tabular*}{\linewidth}{@{\extracolsep{\fill}}lrrrr@{}}
        \toprule
        Repeat
            & Generic Rewrite
            & EBU-Evolve
            & \makecell{Net additional\\correct answers}
            & Gain \\
        \midrule
        R1
            & 74/87 (85.06\%)
            & 82/87 (94.25\%)
            & 8 & +9.20 pp \\
        R2
            & 71/87 (81.61\%)
            & 81/87 (93.10\%)
            & 10 & +11.49 pp \\
        R3
            & 73/87 (83.91\%)
            & 83/87 (95.40\%)
            & 10 & +11.49 pp \\
        \midrule
        Mean
            & 83.52\% & 94.25\% & 9.33 & +10.73 pp \\
        Sample SD
            & 1.76 pp & 1.15 pp & 1.15 & 1.33 pp \\
        \bottomrule
    \end{tabular*}
\end{table}
\FloatBarrier

Across the three independent optimization runs, EBU-Evolve answers 8, 10, and 10 more tasks correctly than Generic Rewrite, averaging 9.33 tasks or 10.73 pp. The per-run advantage ranges from 9.20 to 11.49 pp, so the mean improvement is supported by three comparisons with the same direction. After repeating reflection, candidate generation, and validation selection, the full method retains its advantage across different skill artifacts.

EBU-Evolve accuracy ranges from 93.10\% to 95.40\%, with a sample standard deviation of 1.15 pp; Generic Rewrite ranges from 81.61\% to 85.06\%, with a standard deviation of 1.76 pp. The full method therefore shows both higher mean accuracy and less variation in test performance across these three runs.

\subsubsection{Distribution of Gains across Task Groups}
\label{app:task-group-robustness}

In the single comparison in Appendix~B.8.2, five of the seven net additional correct answers come from Abstract screening. To examine whether gains remain concentrated in that group across optimization repeats, we first report the four groups separately and then recompute performance on the 67 tasks remaining after removing its 20 tasks. Both analyses use the frozen skills and test results from Table~\ref{tab:B9}.

\begin{table}[!htbp]
    \centering\small
    \caption{Correct-answer counts by task group across three independent optimization repeats. Each entry is Generic Rewrite $\rightarrow$ EBU-Evolve. Mean gains are computed within each group's task denominator; the overall row aggregates over all 87 tasks.}
    \label{tab:B10}

    \begin{tabular*}{\linewidth}{@{\extracolsep{\fill}}lrrrrr@{}}
        \toprule
        Task group & Tasks & R1 & R2 & R3 & Mean gain \\
        \midrule
        Abstract screening
            & 20
            & $13\rightarrow18$
            & $12\rightarrow18$
            & $13\rightarrow19$
            & +28.33 pp \\
        Clinical evidence screening
            & 13
            & $12\rightarrow13$
            & $11\rightarrow12$
            & $12\rightarrow13$
            & +7.69 pp \\
        Enrichment review
            & 27
            & $26\rightarrow27$
            & $26\rightarrow26$
            & $25\rightarrow27$
            & +3.70 pp \\
        Variant functional evidence
            & 27
            & $23\rightarrow24$
            & $22\rightarrow25$
            & $23\rightarrow24$
            & +6.17 pp \\
        \midrule
        Overall
            & 87
            & $74\rightarrow82$
            & $71\rightarrow81$
            & $73\rightarrow83$
            & +10.73 pp \\
        \bottomrule
    \end{tabular*}
\end{table}

Abstract screening has the largest improvement, with a mean gain of 28.33 pp (Figure~\ref{fig:components-stability}b). It contributes 5, 6, and 6 net additional correct answers across the three runs and is the main source of the overall advantage. Clinical evidence screening, Variant functional evidence, and Enrichment review also improve, with mean gains of 7.69, 6.17, and 3.70 pp, respectively. Every task-group-by-run net gain is positive or tied, with a tie for Enrichment review in R2, showing that the improvements extend to the other groups.

The gains should also be read alongside the control's accuracy. Generic Rewrite already answers 11--12 of 13 Clinical evidence screening tasks and 25--26 of 27 Enrichment review tasks correctly, leaving less room for further improvement; it answers 12--13 of 20 Abstract screening tasks correctly. A smaller percentage-point gain therefore does not by itself imply that the components have no value for a task group.

To further examine the dependence of the overall advantage on Abstract screening, we exclude this group and recompute the results of each optimization run on the remaining 67 test tasks (Table~\ref{tab:B11}).

\begin{table}[!htbp]
    \centering\small
    \caption{Performance after excluding Abstract screening, leaving 67 test tasks. Each row uses the same frozen skill artifacts as the corresponding 87-task repeat.}
    \label{tab:B11}

    \begin{tabular*}{\linewidth}{@{\extracolsep{\fill}}lrrrr@{}}
        \toprule
        Repeat
            & Generic Rewrite
            & EBU-Evolve
            & \makecell{Net additional\\correct answers}
            & Gain \\
        \midrule
        R1
            & 61/67 (91.04\%)
            & 64/67 (95.52\%)
            & 3 & +4.48 pp \\
        R2
            & 59/67 (88.06\%)
            & 63/67 (94.03\%)
            & 4 & +5.97 pp \\
        R3
            & 60/67 (89.55\%)
            & 64/67 (95.52\%)
            & 4 & +5.97 pp \\
        \midrule
        Mean
            & 89.55\%
            & 95.02\%
            & 3.67
            & +5.47 pp \\
        \bottomrule
    \end{tabular*}
\end{table}
\FloatBarrier

After excluding Abstract screening, EBU-Evolve still answers 3, 4, and 4 more tasks correctly across the three runs, increasing mean accuracy from 89.55\% to 95.02\%, or 5.47 pp. The gain is smaller than the 10.73 pp on all 87 tasks, reflecting the substantial contribution of Abstract screening; EBU-Evolve nevertheless shows improvements in a consistent direction across the other three task groups. EBU-Evolve's gains thus vary by task, while its overall advantage is not wholly determined by a single task group.

These analyses characterize EBU-Evolve's revision quality from complementary perspectives. The component ablations show that paired execution evidence, gain-based case selection, and utilization diagnosis improve skill revision across the tested combinations, with the full method achieving the highest mean accuracy. Independent optimization repeats show that EBU-Evolve's advantage over generic rewriting recurs across different skill artifacts generated and selected in separate runs. The task-group analysis shows that gains are more pronounced for some task types but do not depend entirely on a single group. Together, these experiments further support EBU-Evolve's central design: \textbf{using same-task execution differences to identify utilization problems in existing skills, and translating paired evidence and utilization diagnosis into targeted skill revisions that help agents use existing resources during autonomous execution.} The magnitude of the gains depends on the task type and the existing skill's performance.

\FloatBarrier

\subsubsection{Local Performance Regressions and Source-Cluster Uncertainty}
\label{app:atria-regression}

The preceding analyses examine EBU-Evolve's average advantage over generic rewriting and its performance across optimization repeats. We next examine how aggregate gains coexist with local regressions and how shared task sources affect uncertainty. In the Atria-Dawn-Preview/Claude Code evaluation, correctness increases from 42/87 to 77/87 overall, while Clinical evidence screening decreases from 10/13 to 8/13. The latter includes one repair and three regressions, motivating a task-level account of the net change.

As Atria-Dawn-Preview is no longer available, we use its released version, Atria-Dawn, for the further experiments. We report the frozen Atria-Dawn-Preview evaluation and the three independent optimization runs with Atria-Dawn separately. The former characterizes the Clinical decline; the latter examines the distribution of gains and regressions after reoptimization. All analyses use Natural Skill Use on the same 87 test tasks across four task groups.

\paragraph{First, independent optimization examines aggregate gains and local changes.}
With Atria-Dawn under Claude Code, we fix the development-stage execution evidence, data splits, and test tasks, and independently run the complete optimization procedure three times ($R=3$) from the original skills. Each run repeats reflection, revision, candidate review, and validation selection, followed by separate tests of the original and frozen evolved skills. Runs use independent sessions. In this experiment, each candidate-validation task and each test condition is executed once; the optimizer is Opus4.8, and the task-execution output limit is 64,000 tokens.

\begin{table}[!htbp]
    \centering\small
    \caption{Three independent optimization runs with Atria-Dawn. Original and evolved skills are evaluated in each run; repairs and regressions are paired changes in task correctness.}
    \label{tab:B12}

    \setlength{\tabcolsep}{4pt}
    \begin{tabular*}{\linewidth}{@{\extracolsep{\fill}}llrllr@{}}
        \toprule
        Run
            & \makecell{Overall\\original $\to$ evolved}
            & \makecell{Gain\\(pp)}
            & \makecell{Clinical\\original $\to$ evolved}
            & \makecell{Variant\\original $\to$ evolved}
            & \makecell{Repairs /\\regressions} \\
        \midrule
        Run 1
            & 56/87 $\to$ 81/87
            & +28.74
            & 10/13 $\to$ 13/13
            & 23/27 $\to$ 25/27
            & 26 / 1 \\
        Run 2
            & 59/87 $\to$ 78/87
            & +21.84
            & 12/13 $\to$ 13/13
            & 27/27 $\to$ 22/27
            & 26 / 7 \\
        Run 3
            & 48/87 $\to$ 78/87
            & +34.48
            & 11/13 $\to$ 13/13
            & 24/27 $\to$ 25/27
            & 33 / 3 \\
        \bottomrule
    \end{tabular*}
\end{table}
\FloatBarrier

All three independent optimization runs yield positive aggregate gains, averaging $28.35\pm6.33$ pp (mean $\pm$ sample standard deviation). Evolved Clinical evidence screening reaches 13/13 in every run, and the decline in the Preview evaluation does not recur after these three optimizations. Variant functional evidence decreases from 27/27 to 22/27 in Run 2, with a mean gain of $-2.47$ pp across the three runs. The three runs contain 1, 7, and 3 task-level regressions, respectively. Overall, \textbf{EBU-Evolve's aggregate improvement recurs across independent optimization runs, while local performance changes exhibit differences across optimization runs, task groups, and base-model versions.}

\paragraph{Second, uncertainty is estimated at the source-cluster level.}
A study or original problem can generate multiple task variants. Because some test tasks share an original research question or literature source, we use EBU-Bench construction records to cluster screening tasks by study-source identifier and the other tasks by their original problem \texttt{group\_id}. Abstract screening and Clinical evidence screening contain 18 and eight study clusters, respectively; Enrichment review and Variant functional evidence each contain one original-problem cluster. Deduplicating shared sources across task groups yields 20 clusters in total.

We perform 10,000 bootstrap samples with replacement at the source-cluster level. Whenever a source is sampled, all its tasks and their paired original/evolved outcomes are retained together, including tasks from different groups. We compute the accuracy difference over the sampled tasks and take the 2.5th and 97.5th percentiles as the 95\% interval. Task-group composition can vary with source sampling, allowing the estimate to reflect how different sources contribute to the aggregate gain.

\begin{table}[!htbp]
    \centering\small
    \caption{Exploratory 95\% paired source-cluster bootstrap intervals. Overall estimates resample 20 source clusters; the separate Clinical analysis contains eight clusters. A dash denotes an interval not listed in this table.}
    \label{tab:B13}

    \setlength{\tabcolsep}{4pt}
    \begin{tabular*}{\linewidth}{@{\extracolsep{\fill}}lrrr@{}}
        \toprule
        Evaluation
            & Gain (pp)
            & \makecell{Overall\\95\% CI (pp)}
            & \makecell{Clinical\\95\% CI (pp)} \\
        \midrule
        Atria-Dawn-Preview
            & +40.23 & [14.63, 64.94] & [-38.89, 20.00] \\
        Atria-Dawn, Run 1
            & +28.74 & [10.17, 47.62] & -- \\
        Atria-Dawn, Run 2
            & +21.84 & [-9.76, 54.63] & -- \\
        Atria-Dawn, Run 3
            & +34.48 & [10.46, 60.15] & -- \\
        \bottomrule
    \end{tabular*}
\end{table}
\FloatBarrier

The source-cluster analysis gives an exploratory 95\% confidence interval of \textbf{$[-38.89,20.00]$ pp} for the Clinical evidence screening gain in the initial Atria-Dawn-Preview evaluation. The negative gain is concentrated in two study-source clusters: the four task variants of an acute-gout study (PMID 2454635) decrease from three correct answers to one, and the two variants of a stress-urinary-incontinence study (PMID 17162048) decrease from two correct answers to one.

These three regressions involve the same eligibility boundary: the current tasks admit only randomized controlled trials (RCTs), whereas the study abstracts explicitly describe alternating allocation. All original executions read the design reference and correctly exclude the studies; all evolved executions make no tool calls and incorrectly include them. The local decline is thus concentrated in design-eligibility decisions for a small number of sources.

\paragraph{Third, repairs and regressions are analyzed at the task level.}
We reverify 261 paired original/evolved outcomes across the three source models and count incorrect-to-correct, correct-to-incorrect, both-correct, and both-incorrect transitions. Abstract, Clinical, Enrichment, and Variant in Table~\ref{tab:B14} abbreviate the four task-group names.

\begin{table}[!htbp]
    \centering\small
    \caption{Task-level correctness transitions in the three-source evaluation. Fixed denotes incorrect-to-correct, and regressed denotes correct-to-incorrect; both correct and both wrong denote tasks answered correctly or incorrectly, respectively, in both conditions. Atria here denotes Atria-Dawn-Preview.}
    \label{tab:B14}

    \setlength{\tabcolsep}{3pt}
    \begin{tabular*}{\linewidth}{@{\extracolsep{\fill}}llrrrrrr@{}}
        \toprule
        Model & Task group & Original & Evolved
            & Fixed & Regressed & Both correct & Both wrong \\
        \midrule
        GLM-5.2 & Abstract
            & 8/20 & 20/20 & 12 & 0 & 8 & 0 \\
        GLM-5.2 & Clinical
            & 4/13 & 13/13 & 9 & 0 & 4 & 0 \\
        GLM-5.2 & Enrichment
            & 7/27 & 24/27 & 18 & 1 & 6 & 2 \\
        GLM-5.2 & Variant
            & 17/27 & 27/27 & 10 & 0 & 17 & 0 \\
        \addlinespace
        GPT-5.5 & Abstract
            & 15/20 & 20/20 & 5 & 0 & 15 & 0 \\
        GPT-5.5 & Clinical
            & 7/13 & 13/13 & 6 & 0 & 7 & 0 \\
        GPT-5.5 & Enrichment
            & 10/27 & 27/27 & 17 & 0 & 10 & 0 \\
        GPT-5.5 & Variant
            & 26/27 & 27/27 & 1 & 0 & 26 & 0 \\
        \addlinespace
        Atria & Abstract
            & 10/20 & 19/20 & 9 & 0 & 10 & 1 \\
        Atria & Clinical
            & 10/13 & 8/13 & 1 & 3 & 7 & 2 \\
        Atria & Enrichment
            & 3/27 & 25/27 & 22 & 0 & 3 & 2 \\
        Atria & Variant
            & 19/27 & 25/27 & 7 & 1 & 18 & 1 \\
        \midrule
        GLM-5.2 & Overall
            & 36/87 & 84/87 & 49 & 1 & 35 & 2 \\
        GPT-5.5 & Overall
            & 58/87 & 87/87 & 29 & 0 & 58 & 0 \\
        Atria & Overall
            & 42/87 & 77/87 & 39 & 4 & 38 & 6 \\
        \bottomrule
    \end{tabular*}
\end{table}
\FloatBarrier

Across the three source models, there are 117 repairs and five regressions. Atria-Dawn-Preview's net gain of 35 tasks comprises 39 repairs and four regressions; its net loss of two Clinical evidence screening tasks comprises one repair and three regressions.

\paragraph{Clinical regressions concentrate on study-design eligibility in two sources.}
Further trace auditing shows that the three \textbf{Clinical} regressions involve two worksheet variants of an acute-gout study (PMID 2454635) and one variant of a stress-urinary-incontinence study (PMID 17162048). Both task settings admit only randomized controlled trials, while the abstracts explicitly describe alternating allocation. The original executions read the design reference and correctly exclude the studies; all three evolved executions make no tool calls and incorrectly include them.

\begin{table}[!htbp]
    \centering\small
    \caption{Study-source clusters with changed correctness or persistent errors in Clinical under Atria-Dawn-Preview. Tasks from the other four study clusters remain correct.}
    \label{tab:B15}

    \setlength{\tabcolsep}{4pt}
    \begin{tabular*}{\linewidth}{@{\extracolsep{\fill}}llrlrr@{}}
        \toprule
        PMID
            & Study setting
            & Tasks
            & Correct answers
            & Fixed
            & Regressed \\
        \midrule
        2454635
            & Acute gout
            & 4
            & 3/4 $\to$ 1/4
            & 0 & 2 \\
        17162048
            & Stress urinary incontinence
            & 2
            & 2/2 $\to$ 1/2
            & 0 & 1 \\
        24444360
            & Retained placenta
            & 1
            & 0/1 $\to$ 0/1
            & 0 & 0 \\
        21763393
            & Neonatal resuscitation
            & 2
            & 1/2 $\to$ 2/2
            & 1 & 0 \\
        \bottomrule
    \end{tabular*}
\end{table}
\FloatBarrier

Both task variants of the gout study include a worksheet that recommends including the study and addressing the alternating-allocation issue during quality assessment. However, the current task admits only RCTs, and alternating allocation does not meet this criterion, so the study should be excluded. Even when one worksheet explicitly states that it is unsigned and cannot amend the protocol, the evolved execution still incorrectly includes the study. By comparison, another task variant from the same study with more explicit prompting is answered correctly with both the original and evolved skills, and the evolved execution reads the skill body and references. These results show that variants of the same study can produce different outcomes, with the errors primarily concerning whether the current task's study-design requirements are enforced.

Clinical has a net decrease of two correct answers, comprising three changes: two fewer correct answers for the gout study, one fewer for the incontinence study, and one additional correct answer for the neonatal-resuscitation study. The retained-placenta error persists in both conditions. All three correct-to-incorrect transitions follow the same pattern: the original execution reads the design reference and correctly excludes the study, whereas the evolved execution does not read the skill body and incorrectly includes it.

\paragraph{Regressions in the independent repeats also involve rule application and answer delivery.}
We inspect the five regressions in the three-source evaluation and the eleven regressions across the Atria repeats. Table~\ref{tab:B16} groups the visible execution phenomena.

\begin{table}[!htbp]
    \centering\small
    \caption{Execution phenomena observed in regressed tasks: five events in the three-source evaluation and eleven across the three Atria repeats. A task recurring across runs is counted as a separate execution event.}
    \label{tab:B16}

    \setlength{\tabcolsep}{4pt}
    \begin{tabular*}{\linewidth}{@{\extracolsep{\fill}}lrr@{}}
        \toprule
        Observed execution failure
            & Three-source evaluation
            & Atria repeats \\
        \midrule
        Skill body not read
            & 4 & 6 \\
        Rule or count misapplied after reading
            & 1 & 3 \\
        Correct computation, inconsistent final field
            & 0 & 1 \\
        No answer within the output budget
            & 0 & 1 \\
        \midrule
        Total
            & 5 & 11 \\
        \bottomrule
    \end{tabular*}
\end{table}
\FloatBarrier

The five Variant regressions in Run 2 provide a finer distinction. The original executions read the main file and relevant references in all five tasks; among the evolved executions, two make no tool calls and three read only the main file. All four delivered but incorrect answers retain the correct count of ten independent validation variants, yet assign \texttt{strong} (three tasks) or \texttt{none} (one task) where \texttt{supporting} is required. The error concerns mapping calibration status and validation counts to evidence strength, rather than incorrect calculation of every quantity. The remaining task delivers no final answer within the 64K output budget and receives zero under the evaluation rule.

This run also exposes a specific issue in candidate review and validation selection. The selected body candidate scores 13/13 on validation but loses five correct answers on the test set. Its text generalizes the OddsPath threshold direction as ``greater than,'' whereas the source rules use ``greater than'' for PS3 and ``less than'' for BS3. Review flags this issue in the joint candidate but accepts the body candidate, showing inconsistent review of the same rule across candidates. The test failures also include missing body access and misuse of count-based rules, making both candidate-text checks and execution checks relevant to understanding this decline.

Answer delivery can fail as a separate step. In one Enrichment Run 2 regression, Python correctly computes the probability using $N=12$, $K=3$, and $n=6$, but the final JSON reports $n=3$. In a Variant Run 3 regression, the agent reads the body and all five references yet reports nine validation variants instead of ten. These traces identify three stages at which errors can occur: obtaining resources, applying rules to task conditions, and faithfully carrying computed results into the final answer.

Overall, all three independent Atria optimizations yield substantial positive aggregate gains. Source-cluster analysis further reveals the sensitivity of these gains to source composition, while task-level correctness transitions and execution traces locate local regressions in some task groups within the aggregate gains. These results show that EBU-Evolve improves aggregate autonomous skill-use performance, while the specific gains vary across tasks, sources, and optimized artifacts.

\FloatBarrier

\subsection{Behavioral Cases}

\subsubsection{Rule Not Acquired: Variant Functional Evidence}

On one Variant functional evidence training task, the agent under Natural Skill Use answers after reading only the main SKILL.md and gets it wrong; under Operationally Supported Skill Use, it goes on to read the secondary reference files index, functional, records, frequency, and segregation, and answers correctly. Reflection accordingly proposes a rule that points explicitly to the reference files: the original skill only asked, in general terms, to ``locate the applicable material via the reference index'', whereas the evolved skill requires opening the applicable reference files first, extracting the decision rules, and ``not answering from this SKILL.md alone''.

\subsubsection{Condition Recognized but Not Applied: Enrichment Review}

\begin{figure}[!htb]\centering
\includegraphics[width=\textwidth]{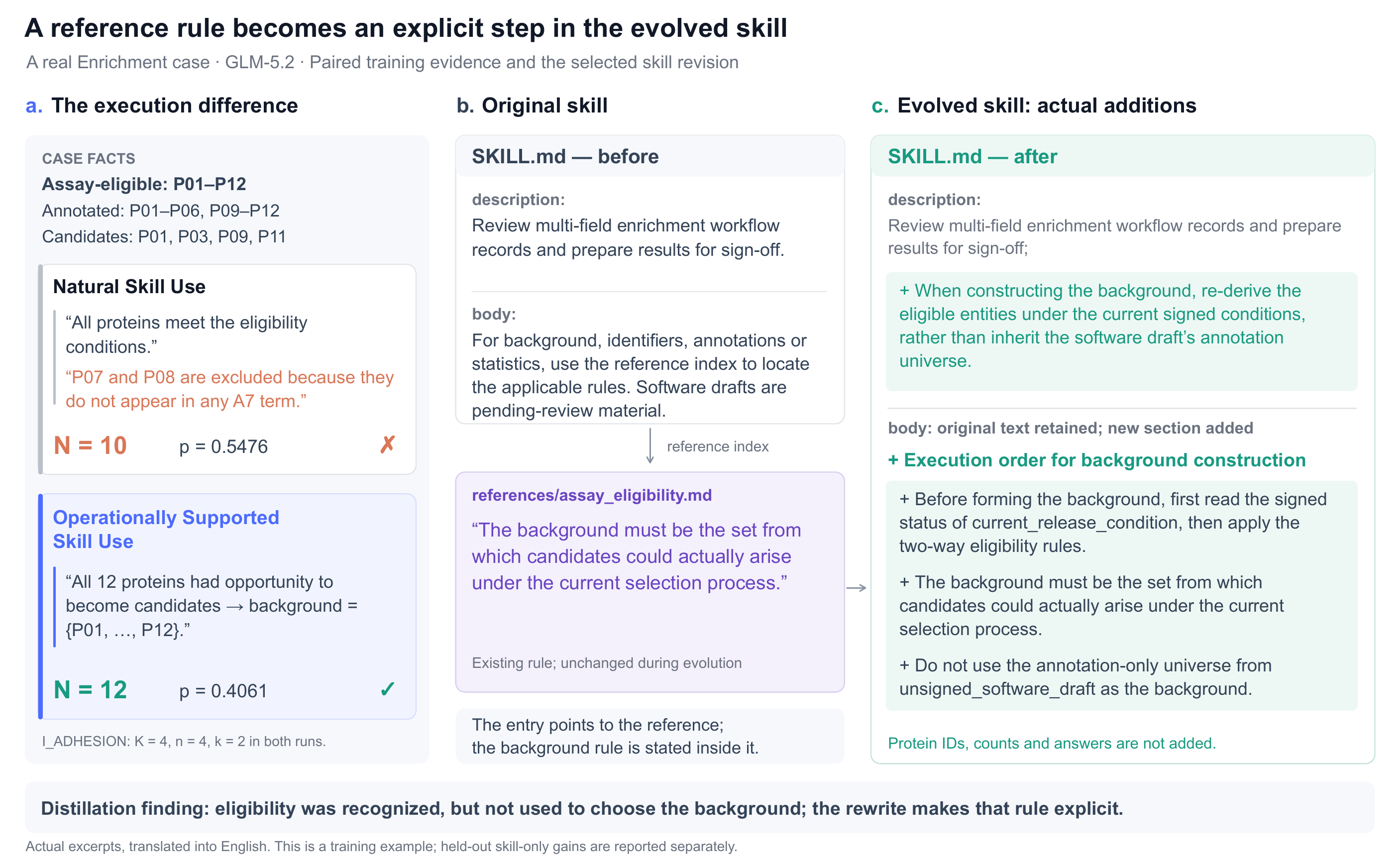}
\caption{Intervention-to-instruction distillation in Enrichment review. The figure shows the executions of one training task under Natural Skill Use and Operationally Supported Skill Use, together with the actual skill revision; the test-task case discussed in the text is a different task, not a re-run of the training task shown here after evolution.}\label{fig:distillation-case}
\end{figure}

Figure~\ref{fig:distillation-case} shows the paired executions of GLM-5.2 on one Enrichment review training task and the corresponding skill revision. Under Natural Skill Use, the agent recognizes 12 assay-eligible entities, yet excludes 2 of them at computation time because they lack term annotations, shrinking the background set to 10; this run reads neither the skill nor the reference materials. Under Operationally Supported Skill Use, these entities are kept and the agent answers correctly. The original skill refers to the software draft as material pending review, and the construction rule for the background set appears only in a reference file; the evolved skill requires rebuilding, before any statistics, the background set of assay-eligible entities under the currently signed-off conditions, rather than inheriting the annotation universe of the draft.

On another Enrichment review test task that took part in neither revision nor version selection, GLM-5.2 reads the same files under the original and the evolved skill: both runs read the main SKILL.md and neither reads the secondary references. The original computes with only the 11 annotated entities (5 candidates) and repeatedly re-checks the statistics without noticing that the object of computation is itself misdefined; the evolved skill keeps all 12 assay-eligible entities (6 candidates) according to the new instruction and answers the task correctly.

\FloatBarrier

\clearpage
% generated by build_en_fable.py from ../en_fable_md/10_appendix_C.md -- do not edit by hand
\section{Evaluation Configuration and Execution Protocol}

\subsection{Model and Harness Configuration}

All evaluations are executed through BenchFlow 0.7.7; each run uses an independent container, with external retrieval and data downloads disabled. The two harness versions are fixed across all runs: Claude Code 2.1.257, and Codex 0.148.0 invoked via codex-acp 1.6.0. Table~\ref{tab:C1} lists the request identifiers of the nine models.

\begin{table}[!htb]
    \centering
    \caption{Model request identifiers.}
    \label{tab:C1}

    \small
    \begin{tabularx}{\linewidth}{
        @{}l
        !{\hspace{5em}}
        >{\raggedright\arraybackslash}X@{}
    }
        \toprule
        Model & Request identifier \\
        \midrule
        GPT-5.5
        & \texttt{gpt-5.5} \\

        Claude Opus 4.8
        & \texttt{claude-opus-4-8} \\

        Gemini 3.8 Flash
        & \texttt{gemini-3.8-flash} \\

        DeepSeek V4 Pro
        & \texttt{deepseek-v4-pro-0813} \\

        GLM-5.2
        & \texttt{glm-5.2}
        (some requests use the alias \texttt{glm-5.2-1m}
        of the same model) \\

        Atria
        & \texttt{Atria-Dawn-Preview} \\

        Kimi K2.7 Code
        & \texttt{kimi-k2.7-code} \\

        Qwen3.8 Max
        & \texttt{qwen3.8-max} \\

        Qwen3.8 27B FP8
        & \texttt{qwen3.8-27b-fp8} \\
        \bottomrule
    \end{tabularx}
\end{table}
\subsection{Scope of Reruns}

The initial time limit is 900 or 3,600 seconds depending on the batch. The 244 evaluation cells that timed out under the 900-second limit (2.2\% of 11,088) were rerun with a budget of up to 7,200 seconds, and the rerun replaced the original record; 55 of them completed after the rerun. If these 55 cells are all counted as failures instead, the mean EBU gap is 43.72 pp, 0.22 pp below the 43.94 pp reported in the main text.

\subsection{Distribution of Invalid Runs}

In the final evaluation, 240 runs (2.2\%) did not receive a valid score and were scored 0: 234 did not submit a valid answer and 6 timed out. The four conditions account for 63, 55, 58, and 64 of these runs, so they are not concentrated in any one condition.

\subsection{Operational-Support Text by Task Source}

Operationally Supported Skill Use reuses the text fixed for each task group. For tasks derived from public benchmarks, the prompt requires the agent to read the associated skill in full, identify the applicable procedure, and apply its steps to the current inputs, for example by extracting and converting calculation parameters or by running code to obtain the required statistics. For tasks authored around research questions, the prompt requires reading the entry file and consulting the necessary reference materials it points to. Neither form of support supplies the answer to the current task, but their requirements differ in specificity, which is why the main text also reports results by task source.

\FloatBarrier

\clearpage
\section{Detailed Evaluation Results}

% generated by build_en_fable.py from ../en_fable_md/11_appendix_D.md -- do not edit by hand
\subsection{Per-Cell Results of the Main Evaluation}

Table~\ref{tab:D1} lists, for nine models under two harnesses and four conditions, the skill-use accuracy, the number of successes, and the EBU gap, corresponding to Figure~\ref{fig:main-results} in the main text.

\begin{table}[!htb]
\centering\small
\caption{Skill-use accuracy and EBU gap of each model–harness combination. Each cell gives "skill-use accuracy (successes/tasks)"; every combination is evaluated on the same 154 tasks, and the EBU gap is in pp. The last two rows are the equally weighted means of the nine models under each harness, which, since the task counts are equal, also equal the pooled successes divided by 1,386. Each combination keeps the score of one run, with invalid runs scored 0 (Appendix C). †: GLM-5.1, used for task screening, belongs to the same model family as GLM-5.2, which may inflate the EBU gap of that family.}\label{tab:D1}
\begin{adjustbox}{max width=\linewidth}
\begin{tabular}{llrrrrr}
\toprule
Model & Harness & \makecell[r]{No\\Skill} & \makecell[r]{Natural\\Skill Use} & \makecell[r]{Prompted\\Skill Use} & \makecell[r]{Operationally\\Supported\\Skill Use} & \makecell[r]{EBU\\gap} \\
\midrule
GPT-5.5 & Claude Code & 20.13 (31/154) & 18.18 (28/154) & 21.43 (33/154) & 80.52 (124/154) & 62.34 \\
GPT-5.5 & Codex & 19.48 (30/154) & 42.86 (66/154) & 39.61 (61/154) & 71.43 (110/154) & 28.57 \\
Claude Opus 4.8 & Claude Code & 18.83 (29/154) & 35.71 (55/154) & 52.60 (81/154) & 87.01 (134/154) & 51.30 \\
Claude Opus 4.8 & Codex & 19.48 (30/154) & 37.66 (58/154) & 46.75 (72/154) & 89.61 (138/154) & 51.95 \\
Gemini 3.8 Flash & Claude Code & 20.78 (32/154) & 35.71 (55/154) & 70.78 (109/154) & 82.47 (127/154) & 46.75 \\
Gemini 3.8 Flash & Codex & 24.03 (37/154) & 68.18 (105/154) & 68.18 (105/154) & 76.62 (118/154) & 8.44 \\
DeepSeek V4 Pro & Claude Code & 23.38 (36/154) & 25.97 (40/154) & 22.73 (35/154) & 72.08 (111/154) & 46.10 \\
DeepSeek V4 Pro & Codex & 22.08 (34/154) & 64.94 (100/154) & 71.43 (110/154) & 79.87 (123/154) & 14.94 \\
GLM-5.2† & Claude Code & 14.94 (23/154) & 15.58 (24/154) & 16.23 (25/154) & 88.96 (137/154) & 73.38 \\
GLM-5.2† & Codex & 16.88 (26/154) & 30.52 (47/154) & 43.51 (67/154) & 92.21 (142/154) & 61.69 \\
Atria & Claude Code & 17.53 (27/154) & 18.83 (29/154) & 18.83 (29/154) & 84.42 (130/154) & 65.58 \\
Atria & Codex & 19.48 (30/154) & 55.19 (85/154) & 60.39 (93/154) & 87.66 (135/154) & 32.47 \\
Kimi K2.7 Code & Claude Code & 22.08 (34/154) & 51.30 (79/154) & 57.79 (89/154) & 77.27 (119/154) & 25.97 \\
Kimi K2.7 Code & Codex & 21.43 (33/154) & 55.19 (85/154) & 58.44 (90/154) & 72.08 (111/154) & 16.88 \\
Qwen3.8 Max & Claude Code & 19.48 (30/154) & 21.43 (33/154) & 25.32 (39/154) & 88.31 (136/154) & 66.88 \\
Qwen3.8 Max & Codex & 20.78 (32/154) & 59.09 (91/154) & 66.88 (103/154) & 81.17 (125/154) & 22.08 \\
Qwen3.8 27B FP8 & Claude Code & 12.99 (20/154) & 15.58 (24/154) & 48.05 (74/154) & 87.01 (134/154) & 71.43 \\
Qwen3.8 27B FP8 & Codex & 14.94 (23/154) & 38.96 (60/154) & 50.65 (78/154) & 83.12 (128/154) & 44.16 \\
Mean over models & Claude Code & 18.90 (262/1,386) & 26.48 (367/1,386) & 37.09 (514/1,386) & 83.12 (1152/1,386) & 56.64 \\
Mean over models & Codex & 19.84 (275/1,386) & 50.29 (697/1,386) & 56.20 (779/1,386) & 81.53 (1130/1,386) & 31.24 \\
\bottomrule
\end{tabular}
\end{adjustbox}
\end{table}

\FloatBarrier

% generated by build_en_fable.py from ../en_fable_md/13_appendix_F.md -- do not edit by hand
\clearpage
\subsection{The Full EBU-Bench}

\begin{figure}[!htb]\centering
\includegraphics[width=\textwidth]{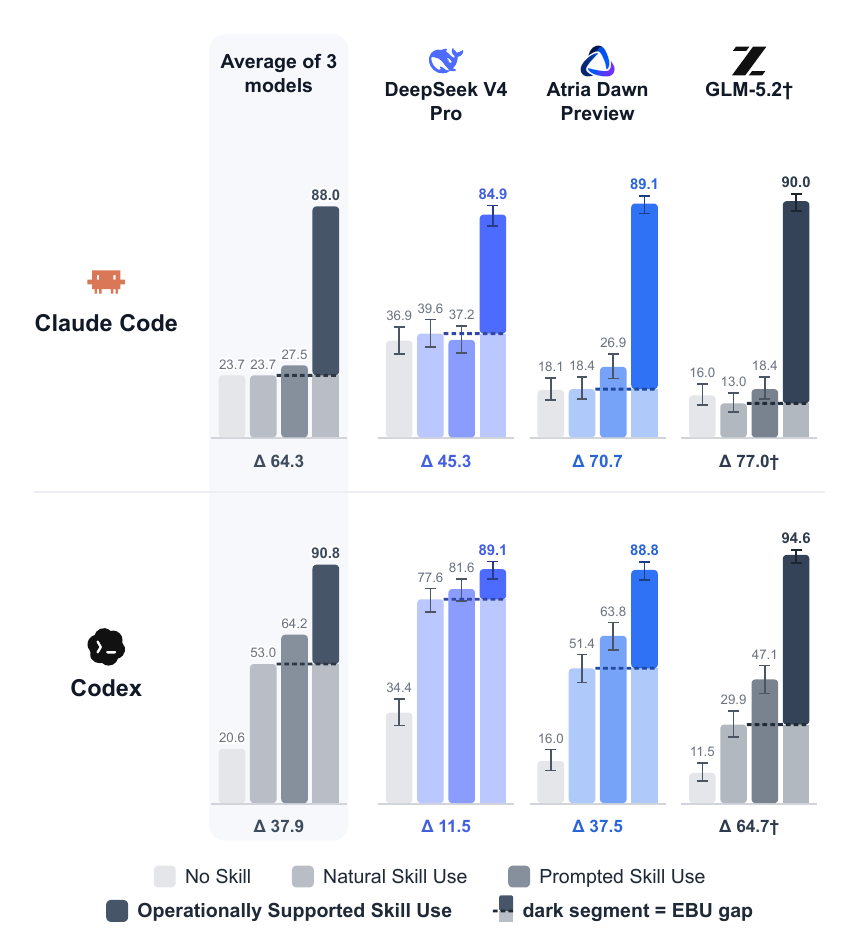}
\caption{Skill-use accuracy (\%) of GLM-5.2, Atria (labeled Atria Dawn Preview in the figure), and DeepSeek V4 Pro under four conditions on the full EBU-Bench (331 tasks). The layout follows the main-results figure: the top row is Claude Code, the bottom row Codex, and the leftmost column the equally weighted mean of the three models; the Operationally Supported Skill Use bar is split at the Natural Skill Use accuracy, the dark upper segment is the EBU gap, and Δ below each bar is its value in pp.~Error bars are per-bar Wilson 95\% intervals (n = 331). †: GLM-5.1, used for task screening, belongs to the same model family as GLM-5.2.}\label{fig:full-benchmark}
\end{figure}

GLM-5.2, Atria, and DeepSeek V4 Pro completed the four-condition evaluation on all 331 tasks of EBU-Bench under Claude Code and Codex (Figure~\ref{fig:full-benchmark}, Table~\ref{tab:D2}). The setup is the same as in Section 4, and the skill-use accuracy of each model--harness combination uses 331 tasks as the denominator; the only difference is that the 177 tasks outside EBU-Bench Lite received no timeout reruns, so timed-out runs are scored 0 directly (86, 25, and 16 runs for GLM-5.2, Atria, and DeepSeek V4 Pro). Even under the least favorable assumption about these timed-out runs, all six combinations retain a positive EBU gap.

The results on the full benchmark agree with EBU-Bench Lite. All six combinations show a positive EBU gap, ranging from 11.48 to 77.04 pp (14.94--73.38 on Lite). Under both harnesses the gaps of the three models are ordered DeepSeek V4 Pro \textless{} Atria \textless{} GLM-5.2, as on Lite, and each model's gap under Claude Code exceeds its gap under Codex. Averaging the three models with equal weight, the EBU gaps of Claude Code and Codex are 64.35 and 37.87 pp.

\begin{table}[!htb]
    \centering
    \setlength{\belowcaptionskip}{6pt}
    \caption{Skill-use accuracy on the full benchmark
    (\%, with successes/tasks in parentheses) and EBU gap (pp).
    The last column is the EBU gap of the same combination on
    EBU-Bench Lite (154 tasks), corresponding to the main results.}
    \label{tab:D2}

    \small
    \setlength{\tabcolsep}{3pt}
    \begin{tabular*}{\linewidth}{
        @{\extracolsep{\fill}}llrrrrrr@{}
    }
        \toprule
        Model
        & Harness
        & No Skill
        & \makecell[r]{Natural\\Skill Use}
        & \makecell[r]{Prompted\\Skill Use}
        & \makecell[r]{Operationally\\Supported\\Skill Use}
        & \makecell[r]{EBU\\gap}
        & \makecell[r]{Lite EBU\\gap} \\
        \midrule

        DeepSeek V4 Pro & Claude Code
        & \makecell[r]{36.86\\(122/331)}
        & \makecell[r]{39.58\\(131/331)}
        & \makecell[r]{37.16\\(123/331)}
        & \makecell[r]{84.89\\(281/331)}
        & 45.32 & 46.10 \\

        DeepSeek V4 Pro & Codex
        & \makecell[r]{34.44\\(114/331)}
        & \makecell[r]{77.64\\(257/331)}
        & \makecell[r]{81.57\\(270/331)}
        & \makecell[r]{89.12\\(295/331)}
        & 11.48 & 14.94 \\

        \addlinespace[3pt]
        Atria & Claude Code
        & \makecell[r]{18.13\\(60/331)}
        & \makecell[r]{18.43\\(61/331)}
        & \makecell[r]{26.89\\(89/331)}
        & \makecell[r]{89.12\\(295/331)}
        & 70.69 & 65.58 \\

        Atria & Codex
        & \makecell[r]{16.01\\(53/331)}
        & \makecell[r]{51.36\\(170/331)}
        & \makecell[r]{63.75\\(211/331)}
        & \makecell[r]{88.82\\(294/331)}
        & 37.46 & 32.47 \\

        \addlinespace[3pt]
        GLM-5.2$^{\dagger}$ & Claude Code
        & \makecell[r]{16.01\\(53/331)}
        & \makecell[r]{12.99\\(43/331)}
        & \makecell[r]{18.43\\(61/331)}
        & \makecell[r]{90.03\\(298/331)}
        & 77.04 & 73.38 \\

        GLM-5.2$^{\dagger}$ & Codex
        & \makecell[r]{11.48\\(38/331)}
        & \makecell[r]{29.91\\(99/331)}
        & \makecell[r]{47.13\\(156/331)}
        & \makecell[r]{94.56\\(313/331)}
        & 64.65 & 61.69 \\

        \bottomrule
    \end{tabular*}
\end{table}
\FloatBarrier

\clearpage
\section{Skill Access and Execution Analysis}

% generated by build_en_fable.py from ../en_fable_md/14_appendix_G.md -- do not edit by hand
\subsection{Counts for Skill-Opening Rates and Paired Analysis}

Table~\ref{tab:E1} gives the counts behind the opening rates and the paired analysis in §5.1. A skill counts as opened once the target skill is successfully read or loaded. The denominator of an opening rate is the number of runs whose opening status can be determined from the execution records.

\begin{table}[!htb]
\centering\small
\setlength{\belowcaptionskip}{6pt}
\caption{Counts for the opening rates and the same-task paired analysis. A pair consists of the Natural Skill Use and Operationally Supported Skill Use runs of the same model, harness, and task; net successes are the difference between the success counts of the two conditions.}\label{tab:E1}
\begin{tabularx}{\linewidth}{Xrr}
\toprule
Statistic & Claude Code & Codex \\
\midrule
Opening rate under Natural Skill Use & 12.45\% (171/1,373) & 65.42\% (906/1,385) \\
Opening rate under Operationally Supported Skill Use & 96.90\% (1,343/1,386) & 99.64\% (1,381/1,386) \\
Net successes corresponding to the EBU gap & 785 & 433 \\
Of which from pairs opened under both conditions & 27 & 182 (42.0\%) \\
Number of pairs opened under both conditions & 169 & 902 \\
Correct answers of these pairs under Natural Skill Use & 117 (69.23\%) & 577 (63.97\%) \\
Correct answers of these pairs under Operationally Supported Skill Use & 144 (85.21\%) & 759 (84.15\%) \\
\bottomrule
\end{tabularx}
\end{table}

Over both harnesses, the net-success increase of Prompted Skill Use relative to Natural Skill Use is 229, which is 18.8\% of the net successes corresponding to the EBU gap (1,218).

In the decomposition of the net EBU gap by opening status (§5.1), a small number of pairs leave the skill unopened under Operationally Supported Skill Use although it was opened under Natural Skill Use; these pairs contribute −0.22 pp in total.

\FloatBarrier

% generated by build_en_fable.py from ../en_fable_md/12_appendix_E.md -- do not edit by hand
\subsection{Runs That Call Unregistered Skill Names}

We count the cases in which the skill name or path requested by the agent is not registered in the current run. Such requests return an ``unknown skill'' error, after which the agent must search again or give up using the skill. The count covers the three conditions that provide the associated skill, 8,316 runs in total (9 models × 2 harnesses × 154 tasks × 3 conditions).

These runs are concentrated in Qwen3.8 Max/Claude Code under Prompted Skill Use (Table~\ref{tab:E2}). In that cell, 60/154 runs call an unregistered skill name: 56 never open the target skill during the whole run yet all submit a valid answer, and the other 4 subsequently open the target skill. The same model and harness show no such calls under Natural Skill Use and 23/154 under Operationally Supported Skill Use, all of which subsequently open the target skill. The Prompted Skill Use row of Figure~\ref{fig:overview}a is one of these cases: the agent calls the unregistered \texttt{acute\_medical\_calculator}, receives ``Unknown skill'', and answers directly without searching for the target skill.

\begin{table}[!htb]
\centering\small
\caption{Number of runs that call an unregistered skill name.}\label{tab:E2}
\begin{tabularx}{\linewidth}{llXXX}
\toprule
Model & Harness & Natural Skill Use & Prompted Skill Use & Operationally Supported Skill Use \\
\midrule
Qwen3.8 Max & Claude Code & 0/154 & 60/154 (56 never open the target skill) & 23/154 (all open the target skill) \\
Qwen3.8 27B FP8 & Claude Code & 0/154 & 3/154 (all open the target skill) & 0/154 \\
Kimi K2.7 Code & Claude Code & 3/154 (all open the target skill) & 2/154 (all open the target skill) & 0/154 \\
Atria & Codex & 1/154 (opens the target skill) & 1/154 (opens the target skill) & 0/154 \\
Other 14 model–harness combinations &  & 0 & 0 & 0 \\
\textbf{Total} &  & \textbf{4/2,772} & \textbf{66/2,772} & \textbf{23/2,772} \\
\bottomrule
\end{tabularx}
\end{table}

Under Prompted Skill Use, the 17 combinations other than Qwen3.8 Max/Claude Code call unregistered skill names in 6/2,618 runs in total, and all 6 subsequently open the target skill.

Across the three conditions there are 110 such requests, involving 88 distinct names or paths. Six are close to the target skill's name (e.g., spelling variants), six are abbreviations of the target skill's category (e.g., \texttt{medcalc}), and the remaining 98 are other unlisted names or paths. The most frequent are \texttt{medcalc} (6 times), \texttt{medical-calculator} and \texttt{multitable-derived-variables-and-association-test} (4 each); the other 85 names each occur 1--2 times.

\FloatBarrier

% generated by build_en_fable.py from ../en_fable_md/15_appendix_H.md -- do not edit by hand
\subsection{A Trace in Which a Memorized Definition Overrides the Specified Rule}

This appendix gives the Atria/Codex trace summarized in §5.1. In the Delta gap task (Na\(^+\) = 130, Cl\(^-\) = 96, HCO\(_3^-\) = 12 mEq/L), the model under Natural Skill Use has read the skill's rule AG − 12 and states that it is ``unambiguous'', yet submits −2, computed from the memorized definition ΔAG − ΔHCO\(_3\). Under Operationally Supported Skill Use, the same model follows the resource, obtaining AG = 130 − (96 + 12) = 22 and hence Delta gap = 10. Against the frozen specification of this task, the two runs are scored as a failure and a success, respectively. This single case cannot establish the overall frequency of such errors; it illustrates the risk that, in medical tasks governed by a specification, a model may read the given rule and still substitute a familiar definition, producing a plausible result that does not comply with the provided specification.

\FloatBarrier

\end{document}